\pdfoutput=1

\documentclass[10pt,twocolumn,letterpaper]{article}

\usepackage[pagenumbers]{wacv} 

\usepackage{amsmath,amssymb}
\usepackage{graphicx}
\usepackage{booktabs}
\usepackage{multirow}
\usepackage{subcaption}
\usepackage{makecell}

\definecolor{wacvblue}{rgb}{0.21,0.49,0.74}
\usepackage[pagebackref,breaklinks,colorlinks,allcolors=wacvblue]{hyperref}

\def\wacvPaperID{2102} 
\def\confName{WACV}
\def\confYear{2027}

\title{Biology-Guided Context-Aware Residual Propagation for Spatial Gene Expression Prediction from Histology}

\author{
Vu Van An\\
Hanoi University of Science and Technology\\
Hanoi, 10000, Vietnam\\
{\tt\small an.vv2416109@sis.hust.edu.vn}
\and
Long D. Nguyen\\
Victoria University of Wellington\\
Wellington, 6012, New Zealand\\
{\tt\small duylong.nguyen@vuw.ac.nz}
\and
Ho Viet Duc Luong\\
Hanoi University of Science and Technology\\
Hanoi, 10000, Vietnam\\
{\tt\small luonghvd@soict.hust.edu.vn}
\and
Binh P. Nguyen\\
Victoria University of Wellington\\
Wellington, 6012, New Zealand\\
{\tt\small binh.p.nguyen@vuw.ac.nz}
}

\begin{document}
\maketitle
\begin{abstract} 
Spatial transcriptomics provides spatially resolved molecular profiling, but its cost, limited throughput and experimental complexity hinder routine large-scale use. Predicting spatial gene expression from hematoxylin-and-eosin-stained histology images offers a scalable alternative, yet remains challenging because histology provides only an indirect morphological view of molecular activity. Existing methods typically decode expression from visual features in a single stage, even when enhanced with spatial aggregation, retrieval or multi-resolution fusion. This formulation is restrictive because spot-level expression reflects not only local morphology but also cellular identity, neighborhood composition and tissue-level organization.
We propose BCARP, a Biology-guided Context-Aware Residual Propagation framework for histology-guided spatial transcriptomics prediction. Instead of directly predicting gene expression from histology embeddings, BCARP treats them as morphology visual priors and progressively refines spot representations in a transcriptomics-oriented latent space. Each refinement step integrates nuclei-guided local cellular context with prototype-guided global context propagation, enabling information exchange across both neighboring and spatially distant spots. A pattern-scale decoder further separates spatial expression into pattern prediction from raw expression with scale calibration. Evaluated on three ST datasets and ten HEST-1k cohorts, BCARP achieves the highest mean correlation-based performance against other baselines. Numerical results support biology-guided local-global residual propagation as an effective alternative to visual decoding for histology-to-transcriptomics prediction.
\end{abstract}

\section{Introduction}
{
Spatial transcriptomics (ST) enables spatially resolved molecular profiling by measuring gene expression while preserving tissue coordinates~\cite{williams2022introduction,gulati2025profiling}. Unlike bulk RNA sequencing, which averages signals over tissue regions, ST quantifies mRNA expression across spatial ``spots'' and reveals how expression patterns are organized within histological architecture. This perspective is critical for biomedical studies, where disease progression, immune response, tumor microenvironment and cell-cell interactions are often spatially structured~\cite{jain2024spatial}, and it bridges conventional histopathology and transcriptomic profiling~\cite{kunhoth2026computational}.}

{
Despite its value, ST remains costly, experimentally demanding and limited in coverage and throughput~\cite{pentimalli2025challenges}, whereas hematoxylin-and-eosin (H\&E)-stained whole-slide images (WSIs) are routinely collected in clinical workflows~\cite{zhang2025deep}. This gap motivates histology-guided ST prediction, which estimates spatial gene expression directly from H\&E images. Beyond patch-based regression such as ST-Net~\cite{he2020integrating}, recent methods enrich the visual representation in complementary ways: graph-based models such as MERGE~\cite{ganguly2025merge} connect spots through spatial and morphological relations; multi-resolution models such as TRIPLEX~\cite{chung2024accurate} and M2OST~\cite{Wang2025m2ost} fuse target-spot, neighborhood and slide-level views; mclSTExp~\cite{min2024mclSTExp} performs contrastive cross-modal alignment for retrieval-based inference; EGN~\cite{yang2024exemplar} refines targets with globally similar exemplars; and STFlow~\cite{huang2025stflow} models the whole-slide expression distribution with flow matching.}

{
Despite this diversity, these approaches largely share a one-stage visual decoding paradigm: contextual information is aggregated into a histology representation, which is then mapped to expression in a single prediction step. This implicitly assumes that the extracted representation is already aligned with the transcriptomic factors underlying spot-level expression. In practice, expression is not determined solely by the appearance of the central patch; it is also shaped by local cellular composition, neighborhood microenvironment and broader tissue organization~\cite{andersson2021spatial,ji2020multimodal,gulati2025profiling}. Histology-guided prediction should therefore go beyond direct visual regression and progressively refine morphology-derived features into transcriptomics-oriented representations.}

To this end, we propose \textbf{BCARP}, a \textbf{B}iology-guided \textbf{C}ontext-\textbf{A}ware \textbf{R}esidual \textbf{P}ropagation framework. BCARP projects frozen pathology foundation embeddings into a task-oriented latent space and treats them as histology anchors. Instead of decoding these anchors in one step, it iteratively refines them through a local-global residual propagation loop: local cellular context encoders use nuclei-derived supervision to inject cell-type and neighborhood-composition information, while a prototype-guided global module combines learnable global priors with spot-level self-attention to model long-range tissue organization. A pattern-scale decoder then separates spatial expression-pattern estimation from raw-expression scale calibration. We evaluate BCARP on HER2ST~\cite{andersson2021spatial}, Skin~\cite{ji2020multimodal} and ST-Net~\cite{he2020integrating} under a unified spot-level protocol, and on ten HEST-1k cohorts~\cite{jaume2024hest} following the STFlow~\cite{huang2025stflow} evaluation suite, spanning multiple organs and cancer types. BCARP outperforms representative baselines, reduces poorly predicted genes, and is further validated by component and refinement-depth ablations.
Our contributions are summarized as follows:
\begin{itemize}

    \item We propose BCARP, a residual propagation framework for multi-step context-aware refinement in histology-guided spatial transcriptomics prediction.

    \item We design a local-global refinement mechanism that combines nuclei-derived local cellular supervision, learnable global context priors and spot-level self-attention to transform histology anchors into transcriptomics-oriented representations.
    \item We conduct comprehensive experiments on three widely used ST datasets and ten curated HEST-1k cohorts, showing that BCARP improves overall cross-cohort performance, achieving the highest mean PCC-M and best mean rank while exhibiting cohort-dependent gains.

\end{itemize}


\section{Related Work}


{Deep learning has become central to WSI analysis, enabling histology images to be represented at cellular, patch and slide levels. Early specialized models focus on local biological structures; for example, HoVer-Net~\cite{graham2019hovernet} performs nuclear instance segmentation and classification by predicting horizontal and vertical distance maps, providing nuclei-level cues for downstream tissue analysis. Beyond individual cells, graph-based approaches such as HEAT~\cite{Chan2023} model WSIs as heterogeneous graphs with nucleus-type node attributes and semantic edge relations, capturing contextual interactions among cellular components. In parallel, self-supervised pathology encoders have been developed to learn transferable patch representations from large unannotated WSI collections. CTransPath~\cite{Wang2022} combines Convolutional Neural Network and Swin Transformer architectures with semantically relevant contrastive learning, while recent large-scale models such as UNI~\cite{chen2024uni}, Phikon-v2~\cite{filiot2024phikon} and Virchow2~\cite{zimmermann2024virchow2} scale DINO-style pretraining to massive pathology datasets and provide strong visual features for diverse diagnostic and biomarker prediction tasks. Other methods further improve WSI-level representation by aggregating patch embeddings into compact morphological prototypes, as in PANTHER~\cite{Song2024} or by aligning histology images with pathology text, as in CONCH~\cite{lu2024conch}. 
}

{
Predicting spatial gene expression from H\&E-stained histology images offers a scalable alternative to experimental spatial transcriptomics. ST-Net~\cite{he2020integrating} formulates the task as patch-level regression, ignoring inter-spot dependencies and broader tissue organization; subsequent work introduces spatial modeling, retrieval, graph construction, multi-scale encoding, multimodal alignment and distribution-level prediction. THItoGene~\cite{Jia2023} models dependencies among spatial locations, while BG-TRIPLEX~\cite{Qu2024} adds boundary, edge and nuclei cues through cross-attention. Among the methods we adopt as baselines, MERGE~\cite{ganguly2025merge} constructs multi-faceted hierarchical graphs that connect spots by both spatial proximity and morphological similarity; TRIPLEX~\cite{chung2024accurate} fuses target-spot, neighborhood and whole-slide views; M2OST~\cite{Wang2025m2ost} regresses a central spot from multi-magnification patches through a many-to-one Transformer, and therefore operates on image crops rather than pre-extracted features; mclSTExp~\cite{min2024mclSTExp} aligns H\&E and spatial-transcriptomic representations by contrastive learning and infers expression through retrieval; EGN~\cite{yang2024exemplar} refines target representations using globally similar exemplars; and STFlow~\cite{huang2025stflow} models the whole-slide joint distribution with flow matching under a zero-inflated negative-binomial prior. Beyond these, MMAP~\cite{Nguyen2026} summarizes slide-level context with a prototype bank, ASIGN~\cite{zhu2025asign} extends imputation to 3D tissue, and GHIST~\cite{Fu2025} and HistoCell~\cite{Zhang2025} predict at cellular resolution.
}

\begin{figure*}[!ht]
    \centering
    \includegraphics[width=1.0\linewidth]{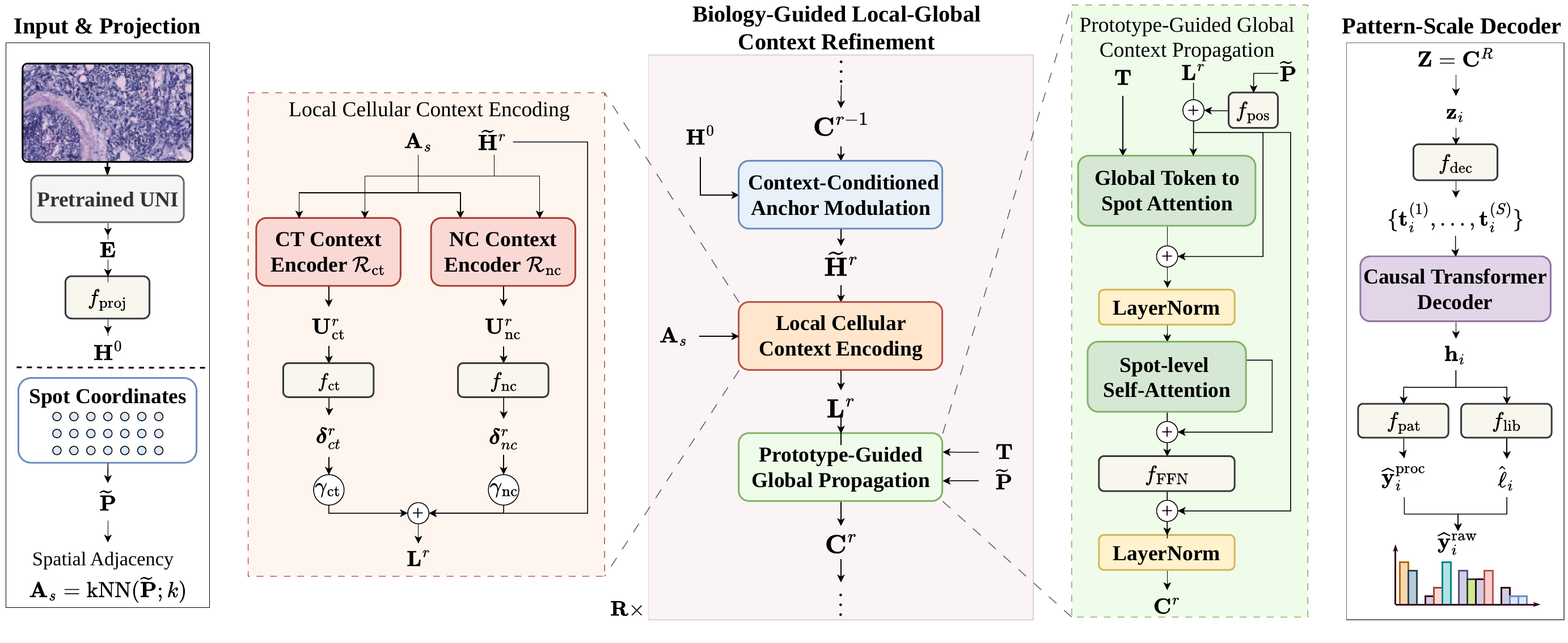}
    \caption{Pipeline overview of BCARP. Frozen histology foundation embeddings are first projected into a shared latent space, then iteratively refined through context-conditioned modulation, local biological encoding and prototype-guided global context propagation. The final representation is passed to a pattern-scale gene expression decoder to generate calibrated spatial gene expression predictions. }
    \label{fig:BCARP_overview}
\end{figure*}


\section{Methodology}
\label{sec:methodology}
{
Figure~\ref{fig:BCARP_overview} illustrates the overall pipeline of BCARP.
Let a tissue section contain $N$ spatial transcriptomics spots. We denote the frozen histology embeddings as $\mathbf{E}\in\mathbb{R}^{N\times d_e}$, the spatial coordinates as $\mathbf{P}\in\mathbb{R}^{N\times 2}$ and the normalized coordinates as $\widetilde{\mathbf{P}}$, obtained by normalizing each coordinate axis by its maximum value. BCARP predicts $G$ target genes, with $\mathbf{Y}^{\mathrm{proc}}\in\mathbb{R}^{N\times G}$ denoting the spatially smoothed target for expression-pattern supervision and $\mathbf{Y}^{\mathrm{raw}}\in\mathbb{R}^{N\times G}$ denoting the raw-count target.
}

\subsection{Histology Embedding Projection and Biology-Guided Local-Global Context Refinement}
\label{sec:embedding_prj}

\paragraph{Histology embedding projection}

{Frozen pathology foundation embeddings provide strong morphology-aware features but are not directly optimized for spatial gene expression prediction. BCARP therefore maps them into a transcriptomics-oriented latent space using a lightweight MLP projector $f_{\mathrm{proj}}(\cdot)$ as
$\mathbf{H}^{0}
=
f_{\mathrm{proj}}
\left(
\mathbf{E}
\right)$, then
$\mathbf{H}^{0}\in\mathbb{R}^{N\times d}$ where $d$ is the hidden dimension. The projected state $\mathbf{H}^{0}$ serves as a stable histology-derived anchor, from which BCARP repeatedly constructs context-conditioned views and refines them with local cellular and global context priors before decoding gene expression.}

\paragraph{Context-aware histology anchor refinement}
{Spatial gene expression is influenced by not only local morphology but also cellular identity, neighborhood composition, and tissue-level organization~\cite{Fu2025}. To capture these dependencies, BCARP performs $R$ context-conditioned refinement steps over $\mathbf{H}^{0}$ using context-conditioned anchor modulation. Let $\mathbf{C}^{r}\in\mathbb{R}^{N\times d}$ denote the context state after step $r$, with $\mathbf{C}^{0}=\mathbf{0}$. At step $r$, we first aggregate the previous context into a section-level summary:
\begin{equation}
\bar{\mathbf{c}}^{r-1}
=
\frac{1}{N}
\sum_{i=1}^{N}
\mathbf{C}^{r-1}_{i},
\qquad
\bar{\mathbf{c}}^{r-1}\in\mathbb{R}^{d}.
\label{eq:aggregate}
\end{equation}
The summary is transformed into feature-wise modulations:
\begin{equation}
\left[
\boldsymbol{\beta}^{r},
\boldsymbol{\gamma}^{r},
\mathbf{a}^{r}
\right]
=
f_{\mathrm{mod}}
\left(
\mathrm{LN}
\left(
\bar{\mathbf{c}}^{r-1}
\right)
\right),
\boldsymbol{\beta}^{r},
\boldsymbol{\gamma}^{r},
\mathbf{a}^{r}
\in\mathbb{R}^{d},
\label{eq1}
\end{equation}
where $\boldsymbol{\beta}^{r}$, $\boldsymbol{\gamma}^{r}$ and $\mathbf{a}^{r}$ denote shift, scale and gate parameters, respectively, $\mathrm{LN}$ denotes LayerNorm, and $f_{\mathrm{mod}}(\cdot)$ is a MLP. These parameters are broadcast to all spots and used to construct a context-modulated histology state:
\begin{equation}
\widetilde{\mathbf{H}}^{r}
=
\mathbf{H}^{0}
+
\tanh
\left(
\mathbf{a}^{r}
\right)
\odot
\left[
\mathrm{LN}
\left(
\mathbf{H}^{0}
\right)
\odot
\left(
1+\boldsymbol{\gamma}^{r}
\right)
+
\boldsymbol{\beta}^{r}
\right],
\label{eq2}
\end{equation}
where $\odot$ denotes element-wise multiplication. This gated modulation preserves the original morphology-derived representation while allowing accumulated context to adaptively reshape it at each refinement step. The resulting state $\widetilde{\mathbf{H}}^{r}$ is then passed to the local and global refinement modules, yielding the update path:
$\mathbf{C}^{r-1}\rightarrow\widetilde{\mathbf{H}}^{r}\rightarrow\mathbf{L}^{r}\rightarrow\mathbf{C}^{r},
\label{eq3}$
where $\mathbf{L}^{r}$ is the locally refined representation obtained from nuclei-guided cellular context encoding and $\mathbf{C}^{r}$ is updated after prototype-guided global context propagation.}

\paragraph{Local cellular context encoding}
{The local stage injects cellular and microenvironmental cues into the context-modulated spot representation. BCARP first constructs a fixed spatial adjacency from normalized spot coordinates by connecting each spot to its $k$ nearest spatial neighbors:
\begin{equation}
\mathbf{A}_{s}
=
\mathrm{kNN}(
\widetilde{\mathbf{P}};k
),
\qquad
\mathbf{A}_{s}\in\{0,1\}^{N\times N}.
\label{eq4}
\end{equation}
The adjacency $\mathbf{A}_{s}$ is computed once and shared across all refinement steps. 
It only specifies the local propagation structure, while the node features are dynamically given by the context-modulated states $\widetilde{\mathbf{H}}^{r}$ at step $r$.
}

{To provide biological supervision, BCARP derives two auxiliary local targets from H\&E-based nuclei analysis. We apply HoVer-Net~\cite{graham2019hovernet} to detect nuclei and assign each nucleus a cell-type pseudo-label. For each spot $i$, we collect nuclei whose centroids fall within a fixed radius around its coordinate $\mathbf{p}_{i}$. Let $T$ denote the number of cell-type categories and let $n_{it}$ be the number of nuclei of type $t$ in the neighborhood of spot $i$. We define the neighborhood-composition target and dominant cell-type target as
\begin{equation}
\mathbf{q}_{i}
=
\left[
\frac{
n_{it}
}{
\sum_{t'=1}^{T} n_{it'}+\epsilon
}
\right]_{t=1}^{T},
\qquad
c_i
=
\arg\max_{t}
n_{it}.
\label{eq5}
\end{equation}

Here $\mathbf{q}_{i}$ represents the local cell-type composition, while $c_i$ summarizes the dominant local phenotype. These targets are used only for training-time auxiliary supervision. 
Given $\widetilde{\mathbf{H}}^{r}$ and $\mathbf{A}_{s}$, BCARP uses two GNN-based local context encoders, indexed by $b\in\{\mathrm{ct},\mathrm{nc}\}$, where $\mathrm{ct}$ is cell-type and $\mathrm{nc}$ denotes neighborhood-composition context:
\begin{equation}
\mathbf{U}_{b}^{r}
=
\mathcal{R}_{b}
\big(
\widetilde{\mathbf{H}}^{r},
\mathbf{A}_{s}
\big),
\qquad
\mathbf{O}_{b}^{r}
=
f_{\mathrm{head},b}
\left(
\mathbf{U}_{b}^{r}
\right),
\label{eq6}
\end{equation}
where $\mathbf{U}_{b}^{r}\in\mathbb{R}^{N\times d_r}$ denotes graph-updated local features, $\mathbf{O}_{b}^{r}\in\mathbb{R}^{N\times T}$ denotes auxiliary logits and $f_{\mathrm{head},b}$ is an MLP. The cell-type branch and neighborhood-composition branch are supervised by $c_i$ and $\mathbf{q}_{i}$, respectively. The corresponding predictions are
\begin{equation}
\widehat{\mathbf{Y}}_{\mathrm{ct}}^{r}
=
\mathrm{softmax}
\left(
\mathbf{O}_{\mathrm{ct}}^{r}
\right),
\qquad
\widehat{\mathbf{Y}}^{r}_{\mathrm{nc}}
=
\mathrm{softmax}
\left(
\mathbf{O}_{\mathrm{nc}}^{r}
\right).
\label{eq7}
\end{equation}
Hidden features from both branches are projected back to the BCARP hidden dimension via an MLP projector $f_{b}$ and injected into the main refinement trajectory:
\begin{equation}
\boldsymbol{\delta}_{b}^{r}
=
f_{b}
\left(
\mathbf{U}_{b}^{r}
\right),
\qquad
\boldsymbol{\delta}_{b}^{r}\in\mathbb{R}^{N\times d},
\qquad
b\in\{\mathrm{ct},\mathrm{nc}\}.
\label{eq8}
\end{equation}
We obtain locally refined state by gated residual injection:
\begin{equation}
\mathbf{L}^{r}
=
\widetilde{\mathbf{H}}^{r}
+
\sum_{b\in\{\mathrm{ct},\mathrm{nc}\}}
\gamma_{b}
\boldsymbol{\delta}_{b}^{r},
\quad
\gamma_{b}=\tanh
\left(
\alpha_{b}
\right),
\label{eq9}
\end{equation}
where $\alpha_{b}$ is a learnable residual-gate parameter for branch $b$. This design allows BCARP to preserve context-modulated histology representation while injecting  the cellular phenotype and the local microenvironmental mixture.
}

\paragraph{Prototype-guided global context propagation}
Local graph encoding captures proximal biological context, but spatial gene expression may depend on broader tissue-level organization. BCARP introduces a small set of learnable global-context prototypes to provide a shared global-context interface. Each spot is allowed to retrieve global prior token information and then exchange the enriched information with other spots through self-attention. We first ground the locally refined representation in tissue geometry. Let $f_{\mathrm{pos}}$ be a spatial positional encoder that maps normalized spot coordinates to $d$-dimensional positional features:
\begin{equation}
\mathbf{X}^{r}
=
\mathbf{L}^{r}
+
f_{\mathrm{pos}}
(
\widetilde{\mathbf{P}}
),
\qquad
\mathbf{X}^{r}\in\mathbb{R}^{N\times d},
\label{eq:global_positional_encoding}
\end{equation}
where $\mathbf{X}^{r}$ is the spatially grounded spot representation at refinement step $r$.
To present the attention operations compactly, we define a multi-head 
cross-attention operator $\mathcal{A}(\mathbf{M},\mathbf{N})$, where 
$\mathbf{M}$ provides queries and $\mathbf{N}$ provides keys and values. 
For the $m$-th attention head, we compute
\begin{equation}
\mathcal{A}_{m}(\mathbf{M},\mathbf{N})
=
\mathrm{softmax}
\left(
\frac{
\mathbf{M}\mathbf{W}^{m}_{Q}
\big(\mathbf{N}\mathbf{W}^{m}_{K}\big)^{\top}
}{
\sqrt{d_h}
}
\right)
\mathbf{N}\mathbf{W}^{m}_{V},
\label{eq:attention_head}
\end{equation}
where $H$ is the number of attention heads and $d_h=d/H$ is the 
dimension of each head. The outputs of all attention heads are 
concatenated to the $d$-dimensional feature:
\begin{equation}
\mathcal{A}(\mathbf{M},\mathbf{N})
=
\mathrm{Concat}
\big(
\mathcal{A}_{1}(\mathbf{M},\mathbf{N}),
\ldots,
\mathcal{A}_{H}(\mathbf{M},\mathbf{N})
\big).
\label{eq:multi_head_attention}
\end{equation}
When $\mathbf{M}\equiv\mathbf{N}$, 
this reduces to standard multi-head self-attention.
BCARP maintains $K_{\mathrm{ctx}}$ learnable global prototype tokens, denoted as $\mathbf{T}\in\mathbb{R}^{K_{\mathrm{ctx}}\times d}$, 
which serve as trainable global context priors without requiring predefined biological classes or physical tissue centroids.
At each refinement step, every spot first attends to these tokens, then fuses the retrieved global context through residual normalization, using global-token-to-spot attention module:
\begin{equation}
\begin{aligned}
\mathbf{O}_{\mathrm{ctx}}^{r}
=
\mathcal{A}
\left(
\mathbf{X}^{r},
\mathbf{T}
\right),
\quad
\bar{\mathbf{X}}_{\mathrm{ctx}}^{r}
=
\mathrm{LN}
\left(
\mathbf{X}^{r}
+
\mathbf{O}_{\mathrm{ctx}}^{r}
\right).
\end{aligned}
\label{eq:token_context_fusion}
\end{equation}
This attention provides each spot with a weighted combination of learned global prototype-context anchors, while the residual normalization injects this global prior into the spatially grounded spot representation before spot-level interaction. 
After receiving prototype-guided global priors, the enriched spot representations exchange and propagate information through spot-level self-attention module:
\begin{equation}
\mathbf{G}^{r}
=
\mathcal{A}
\left(
\bar{\mathbf{X}}_{\mathrm{ctx}}^{r},
\bar{\mathbf{X}}_{\mathrm{ctx}}^{r}
\right),
\qquad
\mathbf{G}^{r}\in\mathbb{R}^{N\times d}.
\label{eq:spot_self_attention}
\end{equation}
We obtain next context state by a residual feed-forward as:
\begin{equation}
\mathbf{C}^{r}
=
\mathrm{LN}
\big(
\mathbf{X}^{r}
+
f_{\mathrm{FFN}}
\big(
\bar{\mathbf{X}}_{\mathrm{ctx}}^{r}
+
\mathbf{G}^{r}
\big)
\big).
\label{eq:context_update}
\end{equation}
The residual path preserves the spatially grounded local state, while the feed-forward branch incorporates both token-routed global context and spot-level attention output. 

\subsection{Pattern-Scale Gene Expression Decoder}
\label{sec:decoder}

{BCARP decodes the final refined spot representation $\mathbf{Z}=\mathbf{C}^{R}\in\mathbb{R}^{N\times d}$ with a pattern-scale gene expression decoder. The decoder factorizes prediction into two complementary components including a pattern branch that estimates relative spatial gene expression structure and a scale branch that calibrates this structure to the raw-count expression space.}

{For each spot $i$, let $\mathbf{z}_{i}\in\mathbb{R}^{d}$ denote the $i$-th row of $\mathbf{Z}$. We first map $\mathbf{z}_{i}$ into the decoder space and expand it into a short sequence of $S$ hierarchy tokens by adding learnable scale embeddings:
\begin{equation}
\mathbf{t}_{i}^{(s)}
=
f_{\mathrm{dec}}
\left(
\mathbf{z}_{i}
\right)
+
\mathbf{e}^{(s)},
\qquad
s=1,\ldots,S.
\label{eq:hierarchy_tokens}
\end{equation}
The scale embeddings induce a compact coarse-to-fine hierarchy over gene groups. A causal transformer decoder processes this sequence, allowing later tokens to condition on earlier decoding states:
\begin{equation}
\mathbf{u}_{i}^{(1)},\ldots,\mathbf{u}_{i}^{(S)}
=
\mathrm{Decoder}
\big(
\mathbf{t}_{i}^{(1)},\ldots,\mathbf{t}_{i}^{(S)}
\big),
\label{eq:causal_decoder}
\end{equation}
where $\mathbf{h}_{i}=\mathbf{u}_{i}^{(S)}$ is used for the spot-level decoding state.} 

\paragraph{Pattern branch}
{Given the decoding state $\mathbf{h}_{i}$, the pattern branch predicts a non-negative processed expression vector:
\begin{equation}
\widehat{\mathbf{y}}_{i}^{\mathrm{proc}}=\widehat{\mathbf{p}}_{i}
=
\operatorname{softplus}
\left(
f_{\mathrm{pat}}
\left(
\mathbf{h}_{i}
\right)
\right),
\qquad
\widehat{\mathbf{p}}_{i}\in\mathbb{R}_{+}^{G}.
\label{eq:pattern_head}
\end{equation}
Stacking the predictions over all spots gives the processed pattern prediction $\widehat{\mathbf{Y}}^{\mathrm{proc}}\in\mathbb{R}_{+}^{N\times G}$. This branch is supervised by the spatially smoothed expression target $\mathbf{Y}^{\mathrm{proc}}$. 
}

\paragraph{Scale calibration branch}

{To calibrate the predicted pattern to the raw-count scale, we normalize $\widehat{\mathbf{p}}_{i}$ into a gene-wise composition $\widehat{\boldsymbol{\pi}}_{i}=\widehat{\mathbf{p}}_{i}/(\mathbf{1}^{\top}\widehat{\mathbf{p}}_{i}+\epsilon)$. Here, $\widehat{\boldsymbol{\pi}}_{i}$ represents the predicted proportion of the total expression assigned to each gene at spot $i$. In parallel, a library-size head predicts a positive spot-level raw-count scale:
\begin{equation}
\widehat{\ell}_{i}
=
\exp
\left(
\operatorname{softplus}
\left(
f_{\mathrm{lib}}
\left(
\mathbf{h}_{i}
\right)
\right)
\right)
-
1,
\qquad
\widehat{\ell}_{i}\in\mathbb{R}_{+}.
\label{eq:library_head}
\end{equation}
The calibrated raw-count prediction is obtained by rescaling the gene-wise composition with the predicted library size:
\begin{equation}
\widehat{\mathbf{y}}_{i}^{\mathrm{raw}}
=
\widehat{\ell}_{i}
\widehat{\boldsymbol{\pi}}_{i}\in\mathbb{R}_{+}^{G}.
\label{eq:raw_prediction}
\end{equation}
Stacking over all spots gives the predicted raw-count matrix $\widehat{\mathbf{Y}}^{\mathrm{raw}}\in\mathbb{R}_{+}^{N\times G}$, where each entry $\widehat{Y}^{\mathrm{raw}}_{ig}$ denotes the predicted raw expression count of gene $g$ at spot $i$.}

{This factorization separates relative gene expression pattern learning from spot-specific raw-count calibration. The pattern branch predicts how expression is distributed across genes, while the scale branch estimates the total count level of each spot and their product yields the final gene-wise raw-count prediction.}

\subsection{Objective}
\label{sec:objective}
\begin{figure*}[!ht]
    \centering
    \includegraphics[width=1\linewidth]{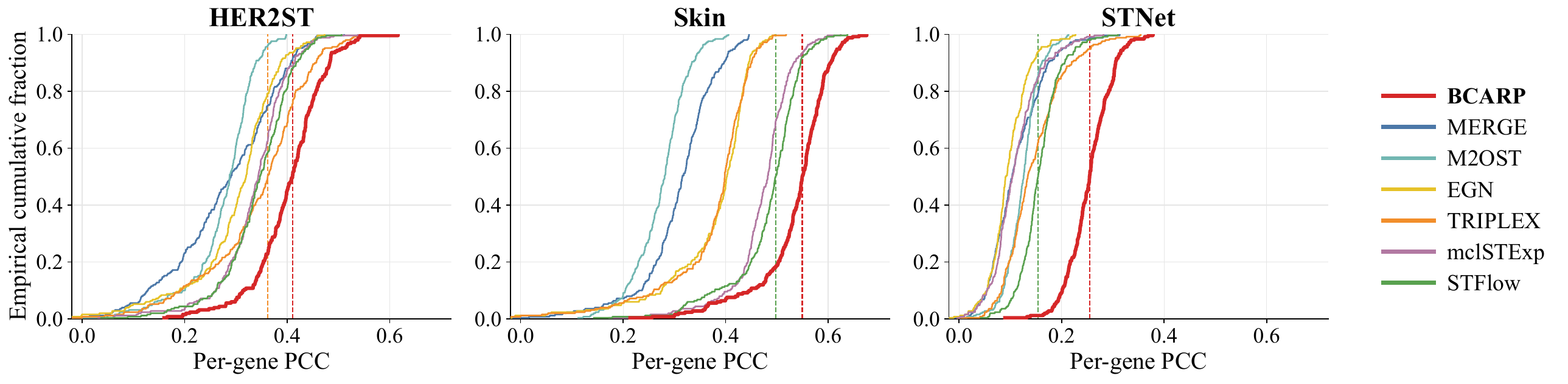}
    \caption{ECDF curves show the distribution of fold-averaged per-gene PCC on HER2ST, Skin and ST-Net datasets. 
    Dashed vertical lines indicate the average per-gene PCC of BCARP and the highest competing baseline.}
    \label{fig:ecdf}
\end{figure*}


BCARP is trained with a multi-task objective that combines processed pattern supervision, raw-count calibration and local biological guidance. 
The processed pattern loss is
$\mathcal{L}_{\mathrm{proc}}
=
\mathcal{L}_{\mathrm{PCC}}
+
\lambda_{\mathrm{huber}}
\mathcal{L}_{\mathrm{Huber}}$,
where
\begin{align}
&\mathcal{L}_{\mathrm{PCC}}
=
1
-
\frac{1}{G}
\sum_{g=1}^{G}
\operatorname{Corr}
\big(
\widehat{\mathbf{Y}}^{\mathrm{proc}}_{:,g},
\mathbf{Y}^{\mathrm{proc}}_{:,g}
\big),\\
&\mathcal{L}_{\mathrm{Huber}}
=
\operatorname{Huber}_{\delta}
\big(
\widehat{\mathbf{Y}}^{\mathrm{proc}},
\mathbf{Y}^{\mathrm{proc}}
\big).
\label{eq:pattern_losses}
\end{align}
For raw-count calibration, $\widehat{\mathbf{Y}}^{\mathrm{raw}}$ is supervised by the raw UMI target $\mathbf{Y}^{\mathrm{raw}}$ using a negative-binomial likelihood:
\begin{equation}
\mathcal{L}_{\mathrm{NB}}
=
-
\frac{1}{NG}
\sum_{i=1}^{N}
\sum_{g=1}^{G}
\log
p_{\mathrm{NB}}
\big(
Y^{\mathrm{raw}}_{ig}
\mid
\widehat{Y}^{\mathrm{raw}}_{ig},
\theta_g
\big),
\label{eq:nb_loss}
\end{equation}
where $\widehat{Y}^{\mathrm{raw}}_{ig}$ is used as the predicted count-rate parameter and $\theta_g$ is the gene-wise dispersion. We further apply a log-space MSE loss to align the calibrated prediction with the raw-count scale:
\begin{equation}
\mathcal{L}_{\mathrm{logMSE}}
=
\operatorname{MSE}
\big(
\log
\big(
1+\widehat{\mathbf{Y}}^{\mathrm{raw}}
\big),
\log
\big(
1+\mathbf{Y}^{\mathrm{raw}}
\big)
\big).
\label{eq:raw_log_mse_loss}
\end{equation}
The decoder loss is
\begin{equation}
\mathcal{L}_{\mathrm{dec}}
=
\mathcal{L}_{\mathrm{proc}}
+
\lambda_{\mathrm{NB}}
\mathcal{L}_{\mathrm{NB}}
+
\lambda_{\mathrm{logMSE}}
\mathcal{L}_{\mathrm{logMSE}}
\label{eq:decoder_loss}
\end{equation}
The local biological branches are supervised by HoVer-Net-derived pseudo-labels. The cell-type loss is averaged over refinement steps:
\begin{equation}
\mathcal{L}_{\mathrm{ct}}
=
\frac{1}{R}
\sum_{r=1}^{R}
\operatorname{CE}
\big(
\widehat{\mathbf{Y}}_{\mathrm{ct}}^{r},
\mathbf{c}
\big),
\label{eq:ct_objective}
\end{equation}
where $\mathbf{c}=\{c_i\}_{i=1}^{N}$ denotes the dominant cell-type pseudo-labels. The neighborhood-composition loss is the average KL divergence between the soft composition target $\mathbf{q}_i$ and the predicted distribution $\widehat{\mathbf{Y}}_{nc,i}^{r}$:
\begin{equation}
\mathcal{L}_{\mathrm{nc}}
=
\frac{1}{R}
\sum_{r=1}^{R}
\frac{1}{N}
\sum_{i=1}^{N}
\sum_{t=1}^{T}
q_{it}
\left[
\log q_{it}
-
\log
\widehat{\mathbf{Y}}^{r}_{nc,it}
\right].
\label{eq:nc_objective}
\end{equation}
The final training objective is
\begin{equation}
\mathcal{L}
=
\mathcal{L}_{\mathrm{dec}}
+
\lambda_{\mathrm{ct}}
\mathcal{L}_{\mathrm{ct}}
+
\lambda_{\mathrm{nc}}
\mathcal{L}_{\mathrm{nc}}.
\label{eq:full_objective}
\end{equation}

\section{Experiments}

\begin{table*}[!ht]
\centering
\small
\caption{Performance comparison on HER2ST, Skin and ST-Net datasets. PCC-H and PCC-M denote average per-gene PCC over highly variable genes and all target genes, respectively. MSE and MAE measure prediction error in expression space, and RVD measures the relative deviation in expression variance between predictions and ground truth. Among the method baselines, best results are shown in bold and second-best results are underlined; the frozen-feature references are reported for context and are excluded from this ranking.}
\label{tab:main_results}
\setlength{\tabcolsep}{1.5pt}
\resizebox{1\textwidth}{!}{
\begin{tabular}{l ccccc c ccccc c ccccc}
\toprule
& \multicolumn{5}{c}{\textbf{HER2ST}} &&
  \multicolumn{5}{c}{\textbf{Skin}} &&
  \multicolumn{5}{c}{\textbf{ST-Net}} \\
\cmidrule{2-6} \cmidrule{8-12} \cmidrule{14-18}
\textbf{Model}
& \textbf{PCC-H}$\uparrow$ & \textbf{PCC-M}$\uparrow$ & \textbf{MSE}$\downarrow$ & \textbf{MAE}$\downarrow$ & \textbf{RVD}$\downarrow$
&& \textbf{PCC-H}$\uparrow$ & \textbf{PCC-M}$\uparrow$ & \textbf{MSE}$\downarrow$ & \textbf{MAE}$\downarrow$ & \textbf{RVD}$\downarrow$
&& \textbf{PCC-H}$\uparrow$ & \textbf{PCC-M}$\uparrow$ & \textbf{MSE}$\downarrow$ & \textbf{MAE}$\downarrow$ & \textbf{RVD}$\downarrow$ \\
\midrule
\multicolumn{18}{l}{\emph{Frozen-feature references}} \\
ResNet50+Ridge
& 0.2249 & 0.2291 & 1.0782 & 0.8224 & 0.9039
&& 0.2695 & 0.3260 & 1.3565 & 0.9050 & 0.8581
&& 0.0985 & 0.0890 & 0.9762 & 0.7862 & 0.9125 \\
UNI+Ridge
& 0.2583 & 0.2598 & 1.0579 & 0.8144 & 0.8840
&& 0.2895 & 0.3262 & 1.3382 & 0.8968 & 0.8812
&& 0.1096 & 0.0936 & 0.9666 & 0.7821 & 0.9223 \\
UNI+MLP
& 0.3474 & 0.3316 & 1.0030 & 0.7890 & 0.8452
&& 0.3029 & 0.3604 & 1.2547 & 0.8670 & 0.8368
&& 0.1843 & 0.1633 & 0.9195 & 0.7609 & 0.8644 \\
UNI+Decoder
& 0.4249 & 0.3721 & 0.9431 & 0.7724 & 0.8188
&& 0.5392 & 0.4930 & 1.4455 & 0.9538 & 1.4511
&& 0.2319 & 0.1868 & 0.8283 & 0.7351 & 0.7685 \\
\midrule
\multicolumn{18}{l}{\emph{Method baselines}} \\
M2OST
& 0.3121 & 0.2756 & 0.9853 & 0.7696 & 2.1442
&& 0.2933 & 0.2739 & 1.3818 & 0.9210 & 0.5739
&& 0.1544 & 0.1239 & 0.7667 & 0.6978 & 1.5675 \\
MERGE
& 0.2913 & 0.2800 & 1.0885 & 0.8274 & 0.8910
&& 0.2827 & 0.3084 & 1.3598 & 0.9052 & 0.9074
&& 0.1214 & 0.1108 & 0.9529 & 0.7776 & 0.9234 \\
EGN
& 0.3884 & 0.2874 & 1.2141 & 0.9953 & 0.8857
&& 0.4039 & 0.3704 & \textbf{0.9505} & 0.8734 & 0.8988
&& 0.1098 & 0.1111 & 0.9826 & 0.7723 & 0.9673 \\
TRIPLEX
& 0.3581 & \underline{0.3427} & 1.0724 & 0.8156 & 0.8990
&& 0.3337 & 0.3749 & 1.3333 & 0.8947 & 0.9137
&& 0.1602 & 0.1450 & 0.9694 & 0.7806 & 0.9263 \\
mclSTExp
& 0.3856 & 0.3356 & 0.9528 & 0.7589 & \underline{0.8143}
&& \underline{0.5212} & 0.4722 & 1.4983 & 0.9862 & \underline{0.4445}
&& 0.1269 & 0.1097 & 0.8659 & 0.7375 & 0.8548 \\
STFlow
& \underline{0.3888} & 0.3410 & \underline{0.8464} & \underline{0.7252} & 0.8555
&& 0.5137 & \underline{0.4832} & 1.1983 & \underline{0.8654} & 0.4951
&& \underline{0.1971} & \underline{0.1565} & \underline{0.7404} & \underline{0.6897} & \textbf{0.7245} \\
\midrule
\textbf{BCARP (Ours)}
& \textbf{0.4567} & \textbf{0.4032} & \textbf{0.8381} & \textbf{0.7191} & \textbf{0.7084}
&& \textbf{0.5852} & \textbf{0.5328} & \underline{1.1818} & \textbf{0.8620} & \textbf{0.2656}
&& \textbf{0.3046} & \textbf{0.2423} & \textbf{0.6761} & \textbf{0.6590} & \underline{0.7500} \\
\bottomrule
\end{tabular}}
\end{table*}

\subsection{Settings}
\label{sec:experimental_settings}


\paragraph{Datasets and preprocessing}
We evaluate BCARP on three established whole-slide spatial transcriptomics benchmarks, HER2ST~\cite{andersson2021spatial}, Skin~\cite{ji2020multimodal}, and ST-Net~\cite{he2020integrating}, as well as ten HEST-1k~\cite{jaume2024hest} cohorts, spanning nine organs and eight cancer types. Each sample consists of an H\&E-stained histology image with spatially resolved gene expression, where each spot is associated with a tissue coordinate, a spot-centered image patch, and an expression vector.
For HER2ST, Skin, and ST-Net, we follow the unified spot-level protocol of MERGE~\cite{ganguly2025merge}: all methods use the same patient-level splits, $224\times224$ spot-centered patches, tissue coordinates, and 250-gene target panel after gene filtering. For HEST-1k, we use the predefined patient-grouped splits for each cohort and select up to 250 highly variable genes, resulting in 250 genes for eight cohorts, 197 genes for LUNG, and 98 genes for PAAD. We apply SPCS smoothing \cite{spcs2022} to obtain processed expression targets for pattern-oriented supervision, while retaining raw-count targets for calibration-oriented objectives. All evaluation metrics are computed in the same $\log(1 + x)$-transformed raw expression space. 

\paragraph{Baselines and implementation details}
We compare BCARP with representative histology-guided expression prediction methods covering graph modeling, multi-resolution fusion, contrastive retrieval, exemplar refinement, whole-slide generative modeling, and frozen-feature prediction. The method baselines include MERGE~\cite{ganguly2025merge}, TRIPLEX~\cite{chung2024accurate}, mclSTExp~\cite{min2024mclSTExp}, EGN~\cite{yang2024exemplar}, STFlow~\cite{huang2025stflow}, and M2OST~\cite{Wang2025m2ost}. For fairness, baselines that support pre-extracted features are evaluated with the same frozen UNI~\cite{chen2024uni} embeddings used by BCARP, replacing their original image encoders. STFlow is already based on UNI features, while M2OST is not encoder-matched because its patch-tokenized multi-field-of-view design requires image crops as input. We further include frozen-feature reference baselines to isolate the effect of the visual backbone and prediction head. On the main datasets, UNI+Ridge, UNI+MLP, and UNI+Decoder evaluate frozen UNI embeddings with increasingly expressive predictors. On HEST-1k, we follow the standard ridge-probe protocol and compare UNI, ResNet50~\cite{He2016}, Virchow2~\cite{zimmermann2024virchow2}, GigaPath~\cite{Xu2024} and CONCH~\cite{lu2024conch} embeddings under the same ridge-regression head. Source code is publicly available at \textcolor{blue}{\url{https://github.com/anvuv/BCARP}}.  

\begin{table*}[!ht]
\centering
\small
\caption{PCC-M performance across the ten HEST-1k cohorts. Mean denotes the average PCC-M across cohorts, and Mean Rank denotes the average within-cohort rank, where lower is better. Best results are shown in bold, and second-best results are underlined.}
\label{tab:cross_dataset}
\resizebox{1\textwidth}{!}{
\begin{tabular}{lcccccccccc@{\hspace{8pt}\vrule\hspace{8pt}}cc}
\toprule
\textbf{Model}
& \textbf{SKCM} & \textbf{LUNG} & \textbf{PAAD} & \textbf{READ} & \textbf{COAD}
& \textbf{LYMPH}
& \textbf{CCRCC} & \textbf{IDC} & \textbf{PRAD} & \textbf{HCC}
& \textbf{Mean} & \makecell{\textbf{Mean}\\ \textbf{Rank} $\downarrow$} \\
\midrule
\multicolumn{13}{l}{\emph{Frozen-feature ridge probes}} \\
ResNet50+Ridge
& 0.5561 & 0.5310 & 0.3963 & 0.0738 & 0.3463 & 0.2642 & 0.1729 & 0.5564 & 0.2005 & 0.0676 & 0.3165 & 10.8 \\
CONCH+Ridge
& 0.6138 & 0.5819 & 0.4585 & 0.1941 & 0.4214 & 0.3131 & 0.1816 & 0.5995 & 0.3031 & 0.0780 & 0.3745 & 7.7 \\
UNI+Ridge
& 0.6340 & 0.5833 & \underline{0.4863} & 0.1987 & 0.4289 & 0.3020 & 0.2077 & \underline{0.6212} & 0.2308 & 0.0868 & 0.3780 & 6.3 \\
GigaPath+Ridge
& 0.6025 & 0.5876 & 0.4823 & 0.2084 & 0.4324 & 0.2958 & 0.1967 & 0.6122 & 0.2886 & 0.0948 & 0.3801 & 6.2 \\
Virchow2+Ridge
& 0.6334 & 0.6005 & 0.4805 & 0.2051 & 0.4220 & 0.3029 & 0.2250 & \textbf{0.6411} & 0.2753 & 0.1118 & 0.3898 & 5.4 \\
\midrule
\multicolumn{13}{l}{\emph{Method baselines}} \\
M2OST
& 0.5651 & 0.5572 & 0.3623 & 0.0666 & 0.0144 & 0.1952 & 0.1149 & 0.4944 & 0.2584 & 0.1155 & 0.2744 & 10.4 \\
MERGE
& 0.5361 & 0.4594 & 0.3751 & 0.1411 & 0.3242 & 0.3120 & 0.2220 & 0.5209 & 0.2759 & 0.1140 & 0.3281 & 9.0 \\
EGN
& 0.6451 & 0.5774 & 0.4586 & 0.1590 & 0.4355 & 0.2420 & 0.1922 & 0.6028 & 0.2299 & 0.0945 & 0.3637 & 7.9 \\
mclSTExp
& 0.5998 & 0.5901 & 0.4603 & 0.2304 & 0.4411 & \underline{0.3243} & 0.2141 & 0.5965 & \underline{0.3296} & 0.1141 & 0.3900 & 4.9 \\
TRIPLEX
& \underline{0.6766} & 0.6042 & 0.4690 & 0.1664 & \underline{0.4415} & 0.3145 & \underline{0.2337} & 0.6157 & 0.3279 & 0.0723 & 0.3922 & 4.3 \\
STFlow
& \textbf{0.6968} & \textbf{0.6315} & \textbf{0.4940} & \textbf{0.2629} & 0.4343 & 0.3134 & 0.2266 & 0.6118 & 0.3153 & \textbf{0.1790} & \underline{0.4166} & \underline{2.6} \\
\midrule
\textbf{BCARP (Ours)}
& 0.6604 & \underline{0.6162} & 0.4857 & \underline{0.2572} & \textbf{0.4491} & \textbf{0.3820} & \textbf{0.2471} & 0.5949 & \textbf{0.3487} & \underline{0.1754} & \textbf{0.4217} & \textbf{2.5} \\
\bottomrule
\end{tabular}}
\end{table*}

\subsection{Comparative Results}
\label{sec:comparative_results}
{
Table~\ref{tab:main_results} reports that BCARP obtains the highest PCC-H and PCC-M on all three datasets.
On HER2ST, BCARP reaches $0.4567/0.4032$, compared with $0.3888/0.3410$ for STFlow and $0.3581/0.3427$ for TRIPLEX, the strongest method baselines on this dataset.
On Skin, BCARP reaches $0.5852/0.5328$ against $0.5137/0.4832$ for STFlow.
The largest improvement is on ST-Net, where BCARP obtains $0.3046/0.2423$ while no competing method exceeds $0.1971/0.1565$.
This indicates that BCARP recovers the spatial layout of expression more faithfully, rather than only reducing pointwise numerical error.
BCARP achieves the lowest MAE across all three datasets, the lowest RVD on HER2ST and Skin, and the lowest MSE on HER2ST and ST-Net.
UNI+Ridge and UNI+MLP remain well below all refinement-based methods, but UNI+Decoder is a strong reference: it reaches $0.4249/0.3721$ on HER2ST and $0.5392/0.4930$ on Skin, which is competitive with, and on Skin better than, every method baseline.
This indicates that a substantial part of the achievable correlation is already carried by frozen pathology embeddings with decoder.
}

{
Table~\ref{tab:cross_dataset} extends the comparison to ten HEST-1k cohorts~\cite{jaume2024hest}, which cover a wider range of organs, cancer types and acquisition platforms.
BCARP obtains the highest mean PCC-M ($0.4217$) and the best mean within-cohort rank ($2.5$), followed by STFlow ($0.4166$, $2.6$).
The two methods are complementary across cohorts: BCARP ranks first on 4/10 cohorts.
The largest single-cohort gain is on LYMPH, where BCARP reaches $0.3820$ against $0.3243$ for the next best method, a cohort in which tissue organization is strongly structured but the local morphological signal is comparatively weak.
We also observe that the ridge probes are competitive in this setting and that Virchow2+Ridge ranks first on IDC ($0.6411$), where BCARP reaches $0.5949$.
This is consistent with the small size of several HEST cohorts, in which a high-capacity refinement model has fewer spots per section to exploit and a linear probe over strong frozen features is a robust alternative.
Overall, the HEST results indicate that the advantage of BCARP transfers beyond the three widely used benchmarks, though the per-cohort margins over STFlow are small and cohort-dependent.
}
\paragraph{Gene-level distribution analysis}
Average correlation can be driven by a small subset of well-predicted genes, so we further examine the full distribution of fold-averaged per-gene PCC using empirical cumulative distribution functions (ECDFs) in Figure~\ref{fig:ecdf}.
A rightward shift of the ECDF indicates that a larger fraction of genes is predicted above any given correlation level.
On all three datasets the BCARP curve lies to the right of every baseline over essentially the entire range, and the shift is most pronounced in the lower half of the distribution: the fraction of genes falling below weak correlation values is markedly reduced.
\begin{figure}
    \centering
    \includegraphics[width=1.0\linewidth]{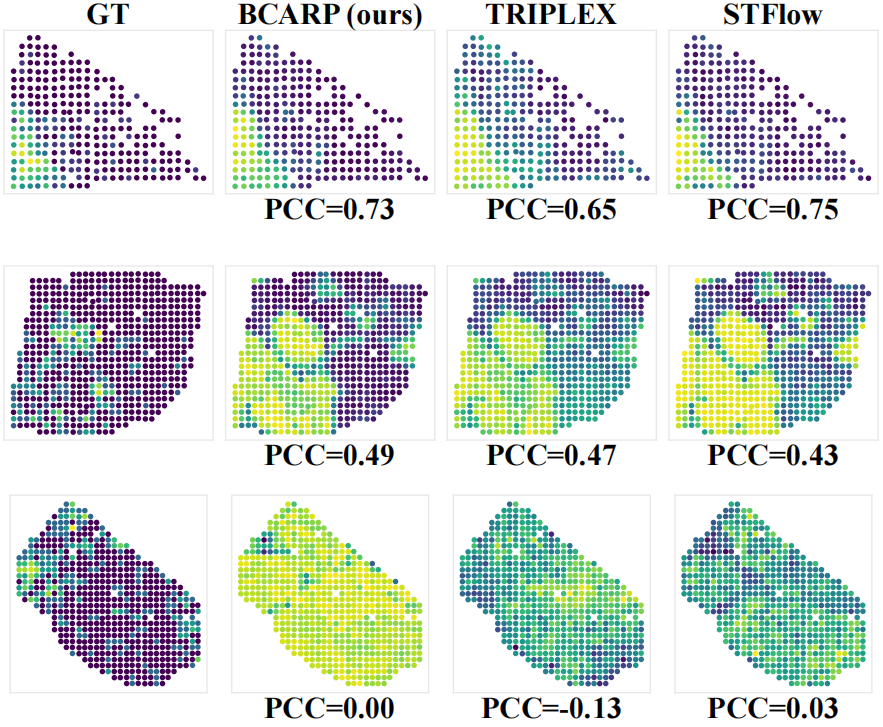}
    \caption{{Qualitative comparison on HER2ST for FN1 on 3 slides.}}
    \label{fig:qualitative}
\end{figure}
{
\paragraph{Qualitative analysis}
Figure~\ref{fig:qualitative} compares predicted and measured FN1 expression on three HER2ST test slides spanning the range of prediction difficulty. On the easiest slide, all methods localise the expressed region and differ only marginally, with STFlow slightly ahead of BCARP. On a moderately difficult slide, BCARP follows the boundary of the expressed region most closely, though all methods overestimate expression across the section and the margin between them is small. On the hardest slide, every method fails, returning nearly inverted or unstructured maps.
}

\begin{table}[!ht]
\centering
\small
\caption{Component ablation on HER2ST. 
}
\label{tab:component_ablation}
\resizebox{.478\textwidth}{!}{
\begin{tabular}{lccccc}
\toprule
Variant & PCC-H$\uparrow$ & PCC-M$\uparrow$ & MSE$\downarrow$ & MAE$\downarrow$ & RVD$\downarrow$ \\
\midrule
Full
& 0.4567 & 0.4032 & \textbf{0.8381} & \textbf{0.7191} & \textbf{0.7084} \\
\midrule
\multicolumn{6}{l}{\emph{Main components}} \\
w/o Proj.
& 0.4283 & 0.3772 & 0.9029 & 0.7512 & 0.8592 \\
w/o CMod
& 0.4379 & 0.3864 & 0.9608 & 0.7751 & 0.7870 \\
w/o Local
& 0.4421 & 0.3882 & 0.8873 & 0.7498 & 0.8138 \\
w/o Global
& 0.4330 & 0.3799 & 0.9264 & 0.7646 & 0.8092 \\
w/o Pos.
& 0.4424 & 0.3884 & 0.8988 & 0.7507 & 0.8285 \\
w/o Dec.
& 0.4571 & 0.4037 & 0.9484 & 0.7736 & 1.6684 \\
\midrule
\multicolumn{6}{l}{\emph{Local-branch ablations}} \\
w/o NC
& \textbf{0.4591} & 0.4041 & 0.8531 & 0.7338 & 0.7552 \\
w/o CT
& 0.4585 & 0.4013 & 0.8941 & 0.7534 & 0.7483 \\
\midrule
\multicolumn{6}{l}{\emph{Global-module ablations}} \\
w/o Prototype
& 0.4586 & \textbf{0.4061} & 0.8538 & 0.7282 & 0.8123 \\
w/o Self-Attn.
& 0.4554 & 0.4032 & 0.8679 & 0.7323 & 0.7724 \\
w/o Global Attn.
& 0.4319 & 0.3803 & 0.8878 & 0.7448 & 0.9665 \\
\bottomrule
\end{tabular}}
\end{table}

\subsection{Ablation Analysis}
\label{sec:ablation_analysis}


\paragraph{Component ablation}
Table~\ref{tab:component_ablation} evaluates the contribution of each BCARP component on HER2ST. We denote the projector, context modulation, local context encoder, global propagation stage, positional encoding, and pattern-scale decoder as Proj., CMod, Local, Global, Pos., and Dec., respectively. For global-module ablations, w/o Global removes the full global stage; w/o Global Attn. removes both prototype attention and spot self-attention but keeps the residual update; w/o Prototype and w/o Self-Attn. remove only the corresponding attention component. The w/o Dec. variant replaces the pattern-scale decoder with a standard MLP head. Removing Proj., CMod, Local, Global, or Pos. consistently weakens performance, showing that BCARP benefits from task-oriented projection, iterative anchor modulation, local biological guidance, long-range context propagation, and explicit tissue geometry. Removing Dec. preserves correlation but substantially worsens MSE, MAE, and RVD, suggesting that the decoder mainly improves expression-scale calibration. 
For the local branches, w/o CT retains only NC, while w/o NC retains
only CT. These results show that neither branch uniformly improves
PCC individually, while the complete local module yields better
expression-error and variance fidelity. The global-module ablations
similarly indicate complementary effects of prototype attention and
spot-level self-attention.


\begin{figure}[t]
    \centering
    \includegraphics[width=1.0\linewidth]{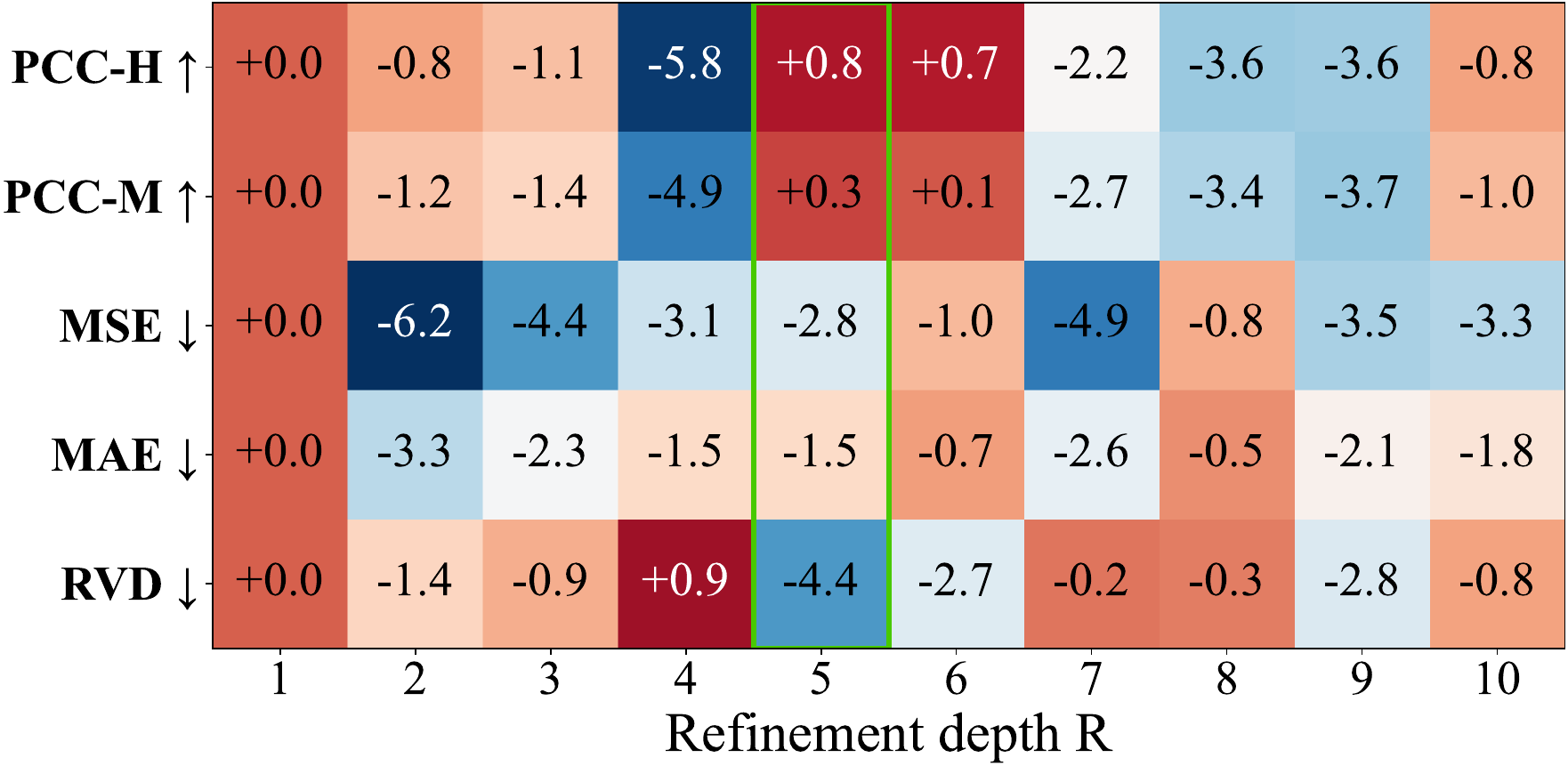}
    \caption{Relative
    percentage change of each metric under different refinement depths compared with $R=1$ on ST-Net dataset. 
    }
    
    \label{fig:r-ablation}
\end{figure} 
\paragraph{Effect of refinement depth}
Figure~\ref{fig:r-ablation} analyzes the effect of refinement depth $R$, with all metrics reported relative to the reference $R=1$ on ST-Net dataset.
Relatively, $R=5$ improves PCC-H/PCC-M by $0.8\%/0.3\%$ and reduces MSE, MAE and RVD by $2.8\%$, $1.5\%$ and $4.4\%$, respectively. This indicates that a small number of refinement steps can already provide strong performance, while moderate additional refinement improves calibration and maintains competitive correlation. We therefore use $R=5$ as default.


\section{Conclusion}
\label{sec:conclusion}
This paper introduced BCARP, a biology-guided context-aware residual propagation framework for histology-guided spatial transcriptomics prediction. BCARP treats frozen pathology embeddings as morphology-derived anchors and progressively refines them with nuclei-guided local cellular context, prototype-guided global propagation, and a pattern-scale decoder. Across HER2ST, Skin, ST-Net, and ten HEST-1k cohorts, BCARP achieves
strong correlation-based performance, including the highest mean
PCC-M across the HEST-1k cohorts. Ablations show that anchor conditioning and local--global propagation
support correlation, while the decoder and nuclei-guided branches
contribute more strongly to scale fidelity.
However, limitations remain: local supervision relies on HoVer-Net
pseudo-labels and may be affected by nuclei detection/classification
errors and cross-tissue domain shift, while morphology-only evidence
limits prediction for genes weakly associated with H\&E appearance.
Future work will study label-noise robustness, complementary molecular supervision, adaptive refinement, and efficient attention.

{
\small
\bibliographystyle{ieeenat_fullname}
\bibliography{main}

@article{he2020integrating,
  title = {Integrating spatial gene expression and breast tumour morphology via deep learning},
  volume = {4},
  ISSN = {2157-846X},
  number = {8},
  journal = {Nature Biomedical Engineering},
  publisher = {Springer Science and Business Media LLC},
  author = {He,  Bryan and Bergenstråhle,  Ludvig and Stenbeck,  Linnea and Abid,  Abubakar and Andersson,  Alma and Borg,  Åke and Maaskola,  Jonas and Lundeberg,  Joakim and Zou,  James},
  year = {2020},
  month = {June},
  pages = {827–834}
}

@inproceedings{jaume2024hest,
    author = {Guillaume Jaume and Paul Doucet and Andrew H. Song and Ming Y. Lu and Cristina Almagro-Perez and Sophia J. Wagner and Anurag J. Vaidya and Richard J. Chen and Drew F. K. Williamson and Ahrong Kim and Faisal Mahmood},
    title = {HEST-1k: A Dataset for Spatial Transcriptomics and Histology Image Analysis},
    booktitle = {Advances in Neural Information Processing Systems},
    year = {2024},
    month = dec,
}

@article{Jia2023,
  title = {THItoGene: a deep learning method for predicting spatial transcriptomics from histological images},
  volume = {25},
  ISSN = {1477-4054},
  number = {1},
  journal = {Briefings in Bioinformatics},
  publisher = {Oxford University Press (OUP)},
  author = {Jia,  Yuran and Liu,  Junliang and Chen,  Li and Zhao,  Tianyi and Wang,  Yadong},
  year = {2023},
  month = Nov 
}

@inproceedings{Qu2024,
  title = {Boundary-Guided Learning for Gene Expression Prediction in Spatial Transcriptomics},
  booktitle = {2024 IEEE International Conference on Bioinformatics and Biomedicine (BIBM)},
  publisher = {IEEE},
  author = {Qu,  Mingcheng and Wu,  Yuncong and Di,  Donglin and Su,  Anyang and Su,  Tonghua and Song,  Yang and Fan,  Lei},
  year = {2024},
  month = Dec,
  pages = {445–450}
}

@inbook{Nguyen2026,
  title = {MMAP: A Multi-Magnification and Prototype-Aware Architecture for Predicting Spatial Gene Expression},
  ISBN = {9789819570843},
  ISSN = {1611-3349},
  booktitle = {PRICAI 2025: Trends in Artificial Intelligence},
  publisher = {Springer Nature Singapore},
  author = {Nguyen,  Hai Dang and Pham,  Nguyen Dang Huy and Nguyen,  The Minh Duc and Nguyen,  Dac Thai and Nguyen,  Hang Thi and Nguyen,  Duong M.},
  year = {2026},
  pages = {381–395}
}

@inproceedings{zhu2025asign,
  title={Asign: an anatomy-aware spatial imputation graphic network for 3d spatial transcriptomics},
  author={Zhu, Junchao and Deng, Ruining and Yao, Tianyuan and Xiong, Juming and Qu, Chongyu and Guo, Junlin and Lu, Siqi and Yin, Mengmeng and Wang, Yu and Zhao, Shilin and others},
  booktitle={Proceedings of the Computer Vision and Pattern Recognition Conference},
  pages={30829--30838},
  year={2025}
}

@article{Zhang2025,
  title = {Systematic inference of super-resolution cell spatial profiles from histology images},
  volume = {16},
  ISSN = {2041-1723},
  number = {1},
  journal = {Nature Communications},
  publisher = {Springer Science and Business Media LLC},
  author = {Zhang,  Peng and Gao,  Chaofei and Zhang,  Zhuoyu and Yuan,  Zhiyuan and Zhang,  Qian and Zhang,  Ping and Du,  Shiyu and Zhou,  Weixun and Li,  Yan and Li,  Shao},
  year = {2025},
  month = Feb 
}

@inproceedings{Chan2023,
  title = {Histopathology Whole Slide Image Analysis with Heterogeneous Graph Representation Learning},
  booktitle = {2023 IEEE/CVF Conference on Computer Vision and Pattern Recognition (CVPR)},
  publisher = {IEEE},
  author = {Chan,  Tsai Hor and Cendra,  Fernando Julio and Ma,  Lan and Yin,  Guosheng and Yu,  Lequan},
  year = {2023},
  month = {June},
  pages = {15661–15670}
}

@article{Wang2022,
  title = {Transformer-based unsupervised contrastive learning for histopathological image classification},
  volume = {81},
  ISSN = {1361-8415},
  journal = {Medical Image Analysis},
  publisher = {Elsevier BV},
  author = {Wang,  Xiyue and Yang,  Sen and Zhang,  Jun and Wang,  Minghui and Zhang,  Jing and Yang,  Wei and Huang,  Junzhou and Han,  Xiao},
  year = {2022},
  month = Oct,
  pages = {102559}
}

@article{filiot2024phikon,
  title={Phikon-v2, a large and public feature extractor for biomarker prediction},
  author={Filiot, Alexandre and Jacob, Paul and Mac Kain, Alice and Saillard, Charlie},
  journal={arXiv preprint arXiv:2409.09173},
  year={2024}
}

@article{zimmermann2024virchow2,
  title={Virchow2: Scaling self-supervised mixed magnification models in pathology},
  author={Zimmermann, Eric and Vorontsov, Eugene and Viret, Julian and Casson, Adam and Zelechowski, Michal and Shaikovski, George and Tenenholtz, Neil and Hall, James and Klimstra, David and Yousfi, Razik and others},
  journal={arXiv preprint arXiv:2408.00738},
  year={2024}
}

@inproceedings{Song2024,
  title = {Morphological Prototyping for Unsupervised Slide Representation Learning in Computational Pathology},
  booktitle = {2024 IEEE/CVF Conference on Computer Vision and Pattern Recognition (CVPR)},
  publisher = {IEEE},
  author = {Song,  Andrew H. and Chen,  Richard J. and Ding,  Tong and Williamson,  Drew F.K. and Jaume,  Guillaume and Mahmood,  Faisal},
  year = {2024},
  month = {June},
  pages = {11566–11578}
}

@article{zhang2025deep,
  title={Deep learning-based histomorphological subtyping and risk stratification of small cell lung cancer from hematoxylin and eosin-stained whole slide images},
  author={Zhang, Yibo and Liu, Shilong and Chen, Jun and Chen, Ruanqi and Yang, Zijian and Sheng, Ruyu and Li, Xin and Wang, Taolue and Liu, Hongyu and Yang, Fan and others},
  journal={Genome Medicine},
  volume={17},
  number={1},
  pages={98},
  year={2025},
  publisher={Springer}
}

@article{pentimalli2025challenges,
  title={Challenges and opportunities in the clinical translation of high-resolution spatial transcriptomics},
  author={Pentimalli, Tancredi Massimo and Karaiskos, Nikos and Rajewsky, Nikolaus},
  journal={Annual Review of Pathology: Mechanisms of Disease},
  volume={20},
  number={1},
  pages={405--432},
  year={2025},
  publisher={Annual Reviews}
}

@article{kunhoth2026computational,
  title={Computational methods for breast cancer molecular profiling using routine histopathology: a review},
  author={Kunhoth, Suchithra and Al-maadeed, Somaya and Akbari, Younes and Al Saady, Rafif Mahmood},
  journal={Archives of Computational Methods in Engineering},
  volume={33},
  number={2},
  pages={2299--2321},
  year={2026},
  publisher={Springer}
}

@article{jain2024spatial,
  title={Spatial transcriptomics in health and disease},
  author={Jain, Sanjay and Eadon, Michael T},
  journal={Nature reviews nephrology},
  volume={20},
  number={10},
  pages={659--671},
  year={2024},
  publisher={Nature Publishing Group UK London}
}

@article{gulati2025profiling,
  title={Profiling cell identity and tissue architecture with single-cell and spatial transcriptomics},
  author={Gulati, Gunsagar S and D’Silva, Jeremy Philip and Liu, Yunhe and Wang, Linghua and Newman, Aaron M},
  journal={Nature Reviews Molecular Cell Biology},
  volume={26},
  number={1},
  pages={11--31},
  year={2025},
  publisher={Nature Publishing Group UK London}
}

@article{williams2022introduction,
  title={An introduction to spatial transcriptomics for biomedical research},
  author={Williams, Cameron G and Lee, Hyun Jae and Asatsuma, Takahiro and Vento-Tormo, Roser and Haque, Ashraful},
  journal={Genome Medicine},
  volume={14},
  number={1},
  pages={68},
  year={2022},
  publisher={Springer}
}

@article{pang2021leveraging,
  title        = {Leveraging information in spatial transcriptomics to predict super-resolution gene expression from histology images in tumors},
  author       = {Pang, Minxing and Su, Kenong and Li, Mingyao},
  journal      = {bioRxiv},
  year         = {2021},
  note         = {Preprint}
}

@inproceedings{xie2023spatially,
  title        = {Spatially Resolved Gene Expression Prediction from Histology Images via Bi-modal Contrastive Learning},
  author       = {Xie, Ronald and Pang, Kuan and Chung, Sai and Perciani, Catia and MacParland, Sonya and Wang, Bo and Bader, Gary},
  booktitle    = {Advances in Neural Information Processing Systems},
  volume       = {36},
  year         = {2023},
}

@inproceedings{yang2024exemplar,
  title = {Exemplar Guided Deep Neural Network for Spatial Transcriptomics Analysis of Gene Expression Prediction},
  booktitle = {2023 IEEE/CVF Winter Conference on Applications of Computer Vision (WACV)},
  publisher = {IEEE},
  author = {Yang,  Yan and Hossain,  Md Zakir and Stone,  Eric A and Rahman,  Shafin},
  year = {2023},
  month = Jan,
  pages = {5028–5037}
}

@article{andersson2021spatial,
  title={Spatial deconvolution of {HER2}-positive breast cancer delineates tumor-associated cell type interactions},
  author={Andersson, Alma and Larsson, Ludvig and Stenbeck, Linnea and Salm{\'e}n, Fredrik and Ehinger, Anna and Wu, Sunny Z and Al-Eryani, Ghamdan and Roden, Daniel and Swarbrick, Alex and Borg, {\AA}ke and others},
  journal={Nature Communications},
  volume={12},
  number={1},
  pages={6012},
  year={2021},
  publisher={Nature Publishing Group UK London}
}

@inproceedings{chung2024accurate,
  title        = {Accurate Spatial Gene Expression Prediction by Integrating Multi-Resolution Features},
  author       = {Chung, Youngmin and Ha, Ji Hun and Im, Kyeong Chan and Lee, Joo Sang},
  booktitle    = {Proceedings of the IEEE/CVF Conference on Computer Vision and Pattern Recognition},
  pages        = {11591--11601},
  year         = {2024},
}

@article{graham2019hovernet,
  title        = {HoVer-Net: Simultaneous Segmentation and Classification of Nuclei in Multi-Tissue Histology Images},
  author       = {Graham, Simon and Vu, Quoc Dang and Raza, Shan E. Ahmed and Azam, Ayesha and Tsang, Yee Wah and Kwak, Jin Tae and Rajpoot, Nasir},
  journal      = {Medical Image Analysis},
  volume       = {58},
  pages        = {101563},
  year         = {2019},
}

@article{chen2024uni,
  title        = {Towards a general-purpose foundation model for computational pathology},
  author       = {Chen, Richard J. and Ding, Tong and Lu, Ming Y. and Williamson, Drew F. K. and Jaume, Guillaume and Song, Andrew H. and Chen, Bowen and Zhang, Andrew and Shao, Daniel and Shaban, Muhammad and Williams, Mane and Oldenburg, Lukas and Weishaupt, Luca L. and Wang, Judy J. and Vaidya, Anurag and Le, Long Phi and Gerber, Georg and Sahai, Sharifa and Williams, Walt and Mahmood, Faisal},
  journal      = {Nature Medicine},
  volume       = {30},
  number       = {3},
  pages        = {850--862},
  year         = {2024},
}

@article{lu2024conch,
  title        = {A visual-language foundation model for computational pathology},
  author       = {Lu, Ming Y. and Chen, Bowen and Williamson, Drew F. K. and Chen, Richard J. and Liang, Ivy and Ding, Tong and Jaume, Guillaume and Odintsov, Igor and Le, Long Phi and Gerber, Georg and Parwani, Anil V. and Zhang, Andrew and Mahmood, Faisal},
  journal      = {Nature Medicine},
  volume       = {30},
  number       = {3},
  pages        = {863--874},
  year         = {2024},
}

@article{Fu2025,
  title = {Spatial gene expression at single-cell resolution from histology using deep learning with GHIST},
  volume = {22},
  ISSN = {1548-7105},
  number = {9},
  journal = {Nature Methods},
  publisher = {Springer Science and Business Media LLC},
  author = {Fu,  Xiaohang and Cao,  Yue and Bian,  Beilei and Wang,  Chuhan and Graham,  Dinny and Pathmanathan,  Nirmala and Patrick,  Ellis and Kim,  Jinman and Yang,  Jean Yee Hwa},
  year = {2025},
  month = {Sept},
  pages = {1900–1910}
}

@InProceedings{ganguly2025merge,
    author    = {Ganguly, Aniruddha and Chatterjee, Debolina and Huang, Wentao and Zhang, Jie and Yurovsky, Alisa and Johnson, Travis Steele and Chen, Chao},
    title     = {{MERGE}: {Multi}-faceted Hierarchical Graph-based {GNN} for Gene Expression Prediction from Whole Slide Histopathology Images},
    booktitle = {Proceedings of the IEEE/CVF Conference on Computer Vision and Pattern Recognition (CVPR)},
    month     = {June},
    year      = {2025},
    pages     = {15611-15620}
}

@article{spcs2022,
    author = {Liu, Yusong and Wang, Tongxin and Duggan, Ben and Sharpnack, Michael and Huang, Kun and Zhang, Jie and Ye, Xiufen and Johnson, Travis S},
    title = {SPCS: a spatial and pattern combined smoothing method for spatial transcriptomic expression},
    journal = {Briefings in Bioinformatics},
    volume = {23},
    number = {3},
    pages = {bbac116},
    year = {2022},
    month = {05},
    issn = {1477-4054},
    eprint = {https://academic.oup.com/bib/article-pdf/23/3/bbac116/43744892/bbac116.pdf},
}

@article{ji2020multimodal,
    author = {Ji, Andrew and Rubin, Adam and Thrane, Kim and Jiang, Sizun and Reynolds, David and others},
    year = {2020},
    month = {06},
    pages = {},
    title = {Multimodal Analysis of Composition and Spatial Architecture in Human Squamous Cell Carcinoma},
    volume = {182},
    journal = {Cell},
}

@inproceedings{
    huang2025stflow,
    title={Scalable Generation of Spatial Transcriptomics from Histology Images via Whole-Slide Flow Matching},
    author={Tinglin Huang and Tianyu Liu and Mehrtash Babadi and Wengong Jin and Rex Ying},
    booktitle={Forty-second International Conference on Machine Learning},
    year={2025},
}

@article{Wang2025m2ost,
  title = {M2OST: Many-to-one Regression for Predicting Spatial Transcriptomics from Digital Pathology Images},
  volume = {39},
  ISSN = {2159-5399},
  number = {7},
  journal = {Proceedings of the AAAI Conference on Artificial Intelligence},
  publisher = {Association for the Advancement of Artificial Intelligence (AAAI)},
  author = {Wang,  Hongyi and Du,  Xiuju and Liu,  Jing and Ouyang,  Shuyi and Chen,  Yen-Wei and Lin,  Lanfen},
  year = {2025},
  month = Apr,
  pages = {7709–7717}
}

@article{min2024mclSTExp,
  title={Multimodal contrastive learning for spatial gene expression prediction using histology images},
  author={Min, Wenwen and Shi, Zhiceng and Zhang, Jun and Wan, Jun and Wang, Changmiao},
  journal={Briefings in Bioinformatics},
  volume = {25},
  number = {6},
  pages = {bbae551},
  year={2024}
}

@article{Xu2024,
  title = {A whole-slide foundation model for digital pathology from real-world data},
  volume = {630},
  ISSN = {1476-4687},
  number = {8015},
  journal = {Nature},
  publisher = {Springer Science and Business Media LLC},
  author = {Xu,  Hanwen and Usuyama,  Naoto and Bagga,  Jaspreet and Zhang,  Sheng and Rao,  Rajesh and Naumann,  Tristan and Wong,  Cliff and Gero,  Zelalem and González,  Javier and Gu,  Yu and Xu,  Yanbo and Wei,  Mu and Wang,  Wenhui and Ma,  Shuming and Wei,  Furu and Yang,  Jianwei and Li,  Chunyuan and Gao,  Jianfeng and Rosemon,  Jaylen and Bower,  Tucker and Lee,  Soohee and Weerasinghe,  Roshanthi and Wright,  Bill J. and Robicsek,  Ari and Piening,  Brian and Bifulco,  Carlo and Wang,  Sheng and Poon,  Hoifung},
  year = {2024},
  month = May,
  pages = {181–188}
}

@inproceedings{He2016,
  title = {Deep Residual Learning for Image Recognition},
  booktitle = {2016 IEEE Conference on Computer Vision and Pattern Recognition (CVPR)},
  publisher = {IEEE},
  author = {He,  Kaiming and Zhang,  Xiangyu and Ren,  Shaoqing and Sun,  Jian},
  year = {2016},
  month = Jun,
  pages = {770–778}
}

@article{CellViT,
    title = {CellViT: Vision Transformers for precise cell segmentation and classification},
    journal = {Medical Image Analysis},
    volume = {94},
    pages = {103143},
    year = {2024},
    issn = {1361-8415},
    doi = {https://doi.org/10.1016/j.media.2024.103143},
    url = {https://www.sciencedirect.com/science/article/pii/S1361841524000689},
    author = {Fabian Hörst and Moritz Rempe and Lukas Heine and Constantin Seibold and Julius Keyl and Giulia Baldini and Selma Ugurel and Jens Siveke and Barbara Grünwald and Jan Egger and Jens Kleesiek},
}

@article{chen2025visual,
  title={A visual--omics foundation model to bridge histopathology with spatial transcriptomics},
  author={Chen, Weiqing and Zhang, Pengzhi and Tran, Tu N and Xiao, Yiwei and Li, Shengyu and Shah, Vrutant V and Cheng, Hao and Brannan, Kristopher W and Youker, Keith and Lai, Li and others},
  journal={Nature Methods},
  pages={1--15},
  year={2025},
  publisher={Nature Publishing Group}
}

@article{Gurarslan2026,
  title = {STORM: spatial transcriptomics optimization by resolution via matrix factorization},
  volume = {27},
  ISSN = {1477-4054},
  number = {3},
  journal = {Briefings in Bioinformatics},
  publisher = {Oxford University Press (OUP)},
  author = {Gurarslan,  Deniz and Camargo,  Oscar and Zeyveli,  Omer and Almalioglu,  Yasin and Li,  Yanjun and Turan,  Mehmet and Kahveci,  Tamer},
  year = {2026},
  month = May 
}
}
\newpage
\appendix

\begin{center}
    \textbf{APPENDICES}
\end{center}

\section{Problem Formulation}
\label{sec:supp_problem}
\begin{figure}
\centering
\includegraphics[width=1.0\linewidth]{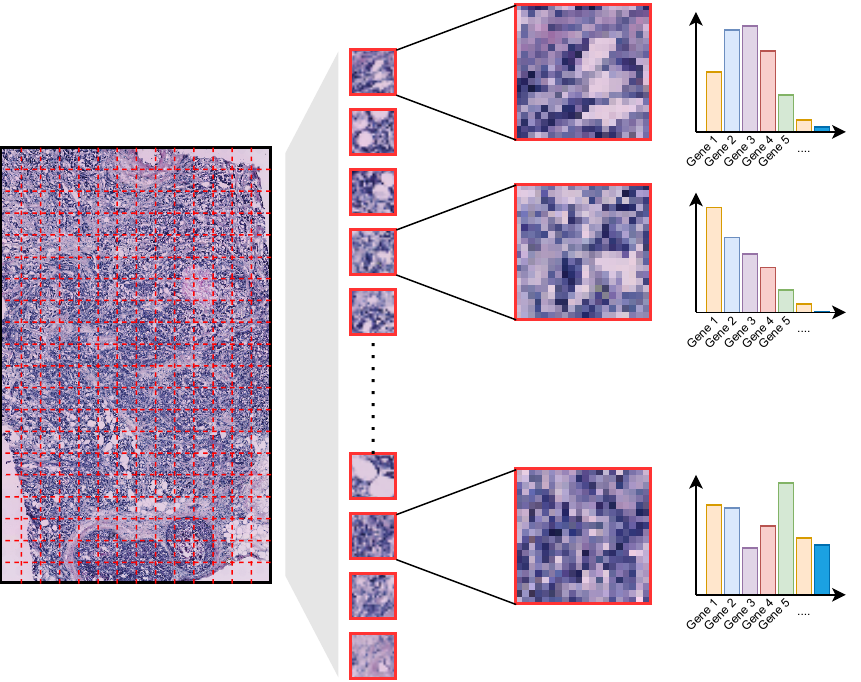}
    \caption{Spatial Transcriptomics Prediction from Histology}
    \label{fig:problem}
\end{figure}

As illustrated in Figure~\ref{fig:problem}, given a whole-slide H\&E image, we represent its spatial-transcriptomics (ST) annotation as a set of spots $\mathcal{S}=\{(\mathbf{x}_i,\mathbf{p}_i,\mathbf{y}_i)\}_{i=1}^{N}$, where $\mathbf{x}_i$ is the spot-centered histology patch, $\mathbf{p}_i\!\in\!\mathbb{R}^{2}$ is its tissue coordinate and $\mathbf{y}_i\!\in\!\mathbb{R}^{G}$ is the measured expression vector over $G$ target genes of spot $i$.
Histology-guided spatial transcriptomics prediction seeks a mapping:
\begin{equation}
F:
\{
\big(
\mathbf{x}_{j},
\mathbf{p}_{j}
\big)\}_{j=1}^{N}
\longmapsto
\widehat{\mathbf{Y}}\in\mathbb{R}^{N\times G},
\qquad
\widehat{\mathbf{y}}_{i}
=
\widehat{\mathbf{Y}}_{i,:}.
\label{eq:supp_task}
\end{equation}
Here, $F$ predicts the section-level expression matrix $\widehat{\mathbf{Y}}$, and its $i$-th row $\widehat{\mathbf{y}}_{i}$ approximates the ground-truth expression profile $\mathbf{y}_{i}$ of spot $i$.

Predicting gene expression strictly from highly localized morphology is inherently ill-posed. Transcriptomic profiling is highly susceptible to technical artifacts, noise and dropout events (where expressed genes are falsely recorded as zeros). Furthermore, the expression profile of a single spot is heavily dependent not only on its own cellular composition but also on the broader tissue microenvironment and intercellular signaling. To address this, BCARP treats frozen pathology-foundation embeddings as morphology-derived anchors. These anchors are progressively refined through a dual mechanism: biology-guided local context encoding and prototype-guided global context propagation, with the aim of improving both local biological consistency and
whole-section spatial context.

\section{Dataset Statistics}
\label{sec:supp_data}
Table~\ref{tab:supp_splits} reports, for every fold, the held-out partition,
the number of training and test slides and the corresponding number of spatial
spots. In total, the evaluation spans $289{,}413$ annotated spots across
thirteen datasets and diverse tissue morphologies.

To prevent patient-level information leakage, all experiments use
patient-disjoint cross-validation, in which every tissue section from a given
patient is assigned exclusively to either the training or the test partition.
HER2ST~\cite{andersson2021spatial} contains eight patients and is split into
eight folds with one held-out patient per fold; Skin~\cite{ji2020multimodal}
contains four patients and is split into four folds, again one patient per
fold; ST-Net~\cite{he2020integrating} contains twenty-three patients, which we
group into eight folds so that each fold holds out two to three patients.
Together these three datasets contribute $52{,}918$ annotated spots.

We additionally evaluate on ten cohorts of the HEST-1k~\cite{jaume2024hest}, spanning skin, lung, pancreas, bowel, lymph
node, kidney, breast, prostate and liver, and acquired with either Visium or
Xenium.
For these cohorts we adopt the official patient-grouped folds released, which adds a further
$236{,}495$ spots and $29$ folds.

\paragraph{Target genes.}

\begin{figure*}
    \centering
    \includegraphics[width=.9\linewidth]{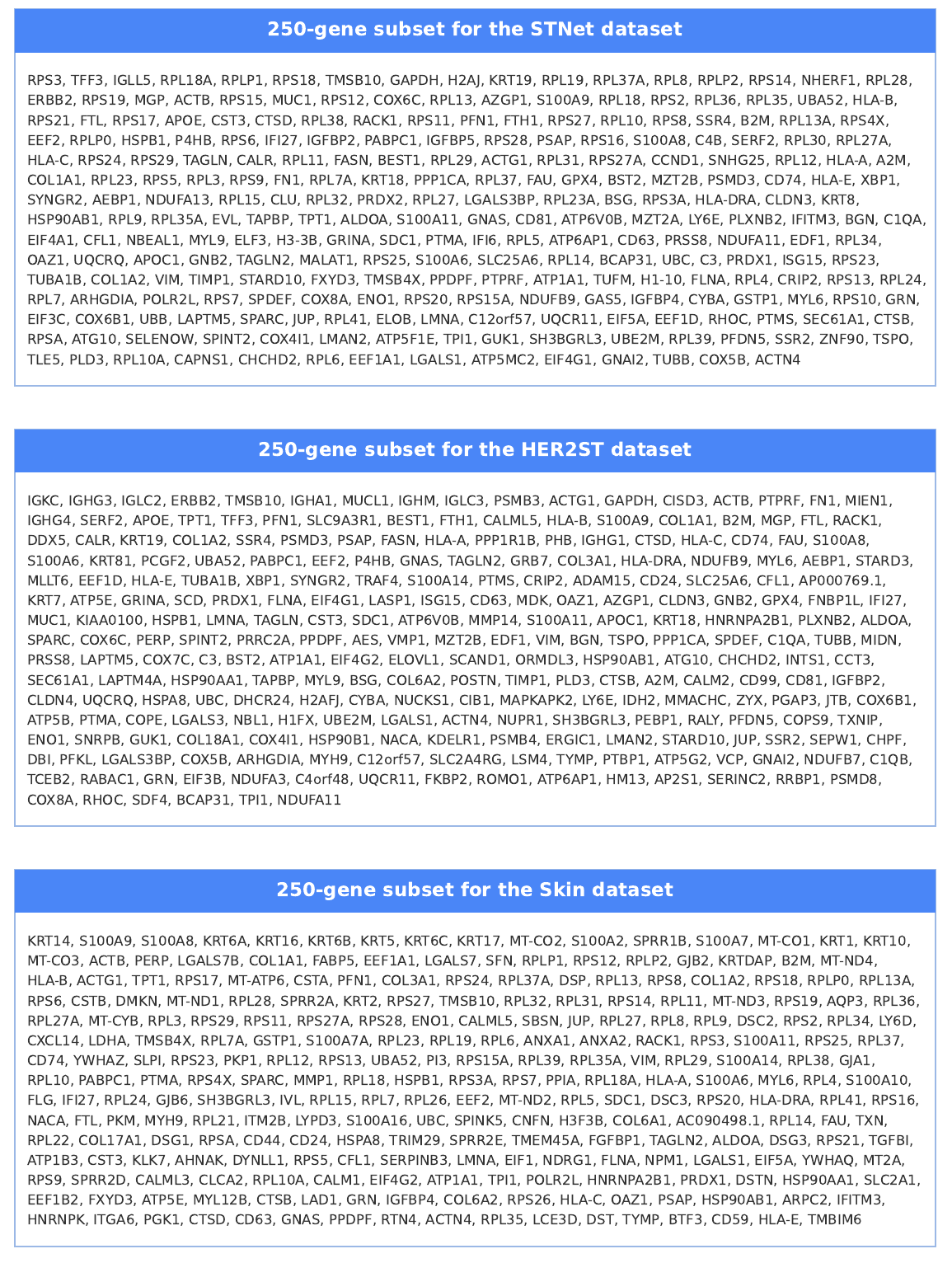}
    \caption{List of the 250 genes used in all experiments on the three ST
    datasets.}
    \label{fig:gene_panel}
\end{figure*}

For HER2ST, Skin, and ST-Net, every method is trained and evaluated
on the same fixed 250-gene panel following the MERGE~\cite{ganguly2025merge}
protocol. This panel corresponds
to the $250$ most-expressed genes per dataset introduced by
TRIPLEX~\cite{chung2024accurate} and reused by MERGE~\cite{ganguly2025merge},
and all metrics are computed in the shared $\log(1+x)$  space.
For the highly variable gene (HVG) analysis, we additionally report results on
a $50$-gene subset per dataset. The HVG subset is selected by a different
criterion, gene-wise variability computed from the training spots only. For
completeness, we list in Figure~\ref{fig:gene_panel} the $250$-gene subsets
used for each ST dataset.
For the HEST-1k cohorts we select up to $250$ highly variable genes per cohort
from the training slides only, following the same criterion as above. The
resulting panel contains $250$ genes for the eight Visium cohorts, but is
capped by the probe set on the two Xenium cohorts, giving $197$ genes for LUNG
and $98$ for PAAD.

\section{Baseline Reproduction}
\label{sec:supp_baselines}

\paragraph{Two reproduction protocols.}
The scores reported in this section are {not} identical to those in Table~1
of the main paper, and the difference is deliberate. In the main paper, every
baseline whose architecture consumes pre-extracted patch features is re-run with
the {same} frozen UNI~\cite{chen2024uni} embeddings as BCARP, so that the
comparison isolates the prediction head from the choice of visual backbone. In
this supplement, we instead retain each baseline's native architecture and visual encoder: every
method keeps the patch encoder, feature dimensionality and architectural choices
prescribed by its original paper or public code, and only the data splits, the
spot-centered patches, the target gene panel and the evaluation metrics are
standardised. 
The main-paper protocol asks how the prediction architectures compare
under a common foundation encoder, whereas the supplementary protocol
evaluates the corresponding native architectures under the same data
and metric setting. The resulting
per-gene correlations differ accordingly, in both directions and by
method-dependent amounts. The two tables should therefore be read side by side.

To ensure a fair comparison, every competing method is reproduced under a single
experimental protocol that follows MERGE~\cite{ganguly2025merge}: identical
patient-level splits, the same spot-centred patches, the same $250$-gene target
panel and the same evaluation metrics. Architectures are re-implemented from
their official publications or public source code whenever available and
hyperparameters follow the original recommendations while sharing a common
optimization protocol. Because BCARP uses frozen UNI embeddings, we
additionally include UNI+MLP and UNI+Decoder probes to separate the contribution
of the refinement loop from that of the pattern-scale decoder. The reproduced
baselines are:
\begin{itemize}\setlength{\itemsep}{1pt}
  \item \textbf{UNI~\cite{chen2024uni}+MLP} -- frozen UNI~\cite{chen2024uni} embeddings are mapped directly
        to gene expression using a standalone MLP predictor.
  \item \textbf{UNI+Decoder} - frozen UNI
        embeddings decoded directly by the pattern-scale decoder, without
        iterative refinement, local biological modelling or global context
        propagation.
  \item \textbf{ST-Net~\cite{he2020integrating}} - a CNN regression framework
        (DenseNet-121 backbone with a linear head) that predicts per-spot
        expression from histology patches without explicit spatial context.
  \item \textbf{HisToGene~\cite{pang2021leveraging}} - a transformer that models dependencies among
        neighbouring spots through spatial attention.
  \item \textbf{BLEEP~\cite{xie2023spatially}} - a contrastive multimodal framework that aligns
        histology and transcriptomics in a shared embedding space; following
        MERGE~\cite{ganguly2025merge}, we use a ResNet18 patch encoder, the same backbone adopted in
        MERGE~\cite{ganguly2025merge} and TRIPLEX~\cite{chung2024accurate}.
  \item \textbf{TRIPLEX~\cite{chung2024accurate}} - a multi-resolution architecture fusing target-patch,
        neighborhood and whole-slide visual information.
  \item \textbf{EGN~\cite{yang2024exemplar}} - an exemplar-guided framework that retrieves similar
        transcriptomic patterns from reference samples to refine predictions.
  \item \textbf{MERGE~\cite{ganguly2025merge}} - a hierarchical graph neural network with spatial
        graph reasoning and multi-scale tissue aggregation, reproduced from its
        official source code.
  \item \textbf{M2OST~\cite{Wang2025m2ost}} - a many-to-one regression transformer that jointly
        consumes patches from several magnification levels of the WSI pyramid to
        predict the expression of their common spot.
  \item \textbf{STFlow~\cite{huang2025stflow}} - a flow-matching model that generates the section-level
        expression matrix through a small number of iterative transport steps.
  \item \textbf{mclSTExp~\cite{min2024mclSTExp}} - a contrastive method that retrieves expression
        profiles from a training bank of spot embeddings.
\end{itemize}
We standardize data splits, patches, target genes and evaluation metrics across
all methods. Training settings are kept as consistent as possible while
respecting method-specific architectural requirements.

\paragraph{Frozen-feature ridge probes.}
On the HEST-1k cohorts the frozen-feature references reported in the main paper
follow the ridge-probe protocol released with the
benchmark~\cite{jaume2024hest}: patch embeddings are extracted once per spot
with the corresponding foundation encoder and regressed onto the target panel
with an $\ell_2$-regularised linear head, using the benchmark's default
regularisation strength rather than a per-cohort tuned value. All encoders are
compared under exactly this protocol, so differences between them reflect the
embeddings and not the head.

\paragraph{On the reproduction of MERGE.}
The MERGE results in our evaluation are not directly comparable to
its published numbers because the protocols differ along several
axes. The original MERGE experiments train and evaluate on
SPCS-smoothed expression using eight-fold cross-validation across
tissue sections. In contrast, our evaluation uses patient-disjoint
folds and computes every reported metric against the raw expression
target after the $\log(1+x)$ transform; SPCS is retained only as
training-time pattern supervision. In addition, the encoder-controlled
results in Table~1 of the main paper replace supported visual encoders
with frozen UNI features. This is particularly relevant to MERGE
because its hierarchical graph includes feature-space clustering, so
changing the visual representation may also alter graph construction.
To separate these effects, Supplementary Table~\ref{tab:main_results2} retains each
baseline's native architecture and encoder while using our
patient-disjoint splits, target panel, and raw-log evaluation. The
main-paper table additionally controls the visual backbone using UNI
where the architecture permits. The resulting MERGE scores should
therefore be interpreted as protocol-specific reproductions rather
than direct replications or contradictions of its published results.

\begin{table*}[t]
\centering
\small
\caption{Performance comparison across the three ST datasets under their native architectural and encoder configurations. Bold indicates the best result and underlining indicates the
second-best result; frozen-feature references are excluded from this
ranking.}
\label{tab:main_results2}
\setlength{\tabcolsep}{1.8pt}
\resizebox{\linewidth}{!}{
\begin{tabular}{l ccccc c ccccc c ccccc}
\toprule
& \multicolumn{5}{c}{\textbf{HER2ST~\cite{andersson2021spatial}}} && \multicolumn{5}{c}{\textbf{Skin~\cite{ji2020multimodal}}} && \multicolumn{5}{c}{\textbf{ST-Net~\cite{he2020integrating}}} \\
\cmidrule{2-6} \cmidrule{8-12} \cmidrule{14-18}
\textbf{Model} & \textbf{PCC-H}\(\uparrow\) & \textbf{PCC-M}\(\uparrow\) & \textbf{MSE}\(\downarrow\) & \textbf{MAE}\(\downarrow\) & \textbf{RVD}\(\downarrow\) && \textbf{PCC-H}\(\uparrow\) & \textbf{PCC-M}\(\uparrow\) & \textbf{MSE}\(\downarrow\) & \textbf{MAE}\(\downarrow\) & \textbf{RVD}\(\downarrow\) && \textbf{PCC-H}\(\uparrow\) & \textbf{PCC-M}\(\uparrow\) & \textbf{MSE}\(\downarrow\) & \textbf{MAE}\(\downarrow\) & \textbf{RVD}\(\downarrow\) \\
\midrule
\multicolumn{18}{l}{\emph{Frozen-feature reference models}} \\
ResNet50+Ridge
& 0.2249 & 0.2291 & 1.0782 & 0.8224 & 0.9039
&& 0.2695 & 0.3260 & 1.3565 & 0.9050 & 0.8581
&& 0.0985 & 0.0890 & 0.9762 & 0.7862 & 0.9125 \\
UNI+Ridge
& 0.2583 & 0.2598 & 1.0579 & 0.8144 & 0.8840
&& 0.2895 & 0.3262 & 1.3382 & 0.8968 & 0.8812
&& 0.1096 & 0.0936 & 0.9666 & 0.7821 & 0.9223 \\
UNI+MLP
& 0.3474 & 0.3316 & 1.0030 & 0.7890 & 0.8452
&& 0.3029 & 0.3604 & 1.2547 & 0.8670 & 0.8368
&& 0.1843 & 0.1633 & 0.9195 & 0.7609 & 0.8644 \\
UNI+Decoder
& 0.4249 & 0.3721 & 0.9431 & 0.7724 & 0.8188
&& 0.5392 & 0.4930 & 1.4455 & 0.9538 & 1.4511
&& 0.2319 & 0.1868 & 0.8283 & 0.7351 & 0.7685 \\
\midrule
\multicolumn{18}{l}{\emph{Method baselines (native configurations)}} \\
ST-Net~\cite{he2020integrating}
& 0.3239 & 0.2430 & 1.2587 & 1.0033 & 2.8133
&& 0.3787 & 0.3352 & 1.0002 & 0.8652 & 3.3435
&& 0.1730 & 0.1509 & 0.7911 & 0.7141 & 0.8978 \\
HisToGene~\cite{pang2021leveraging}
& 0.2882 & 0.2058 & 1.3942 & 1.0733 & 1.1019
&& 0.3076 & 0.2429 & \underline{0.9210} & \underline{0.8270} & 0.7557
&& 0.1149 & 0.1090 & 0.8884 & 0.7562 & 0.8821 \\
BLEEP~\cite{xie2023spatially}
& 0.2846 & 0.2129 & \underline{0.8423} & 0.7962 & 1.6538
&& 0.3694 & 0.3044 & \textbf{0.8024} & \textbf{0.7678} & 1.8789
&& 0.0430 & 0.0351 & 0.8572 & 0.7396 & 3.0336 \\
EGN~\cite{yang2024exemplar}
& 0.2698 & 0.2749 & 1.3371 & 0.8856 & 0.8857
&& 0.2655 & 0.3154 & 1.5425 & 0.9433 & 0.8988
&& 0.0967 & 0.0817 & 1.1714 & 0.8462 & 0.9673 \\
TRIPLEX~\cite{chung2024accurate}
& 0.3581 & \underline{0.3427} & 1.0724 & 0.8156 & 0.8990
&& 0.3337 & 0.3749 & 1.3333 & 0.8947 & 0.9137
&& 0.1602 & 0.1450 & 0.9694 & 0.7806 & 0.9263 \\
mclSTExp~\cite{min2024mclSTExp}
& 0.3627 & 0.3133 & 1.0260 & 0.7867 & 4.8984
&& 0.4675 & 0.4253 & 1.9257 & 1.1175 & 1.1858
&& 0.1451 & 0.1177 & 0.8565 & 0.7327 & 5.6416 \\
M2OST~\cite{Wang2025m2ost}
& 0.3121 & 0.2756 & 0.9853 & 0.7696 & 2.1442
&& 0.2933 & 0.2739 & 1.3818 & 0.9210 & 0.5739
&& 0.1544 & 0.1239 & 0.7667 & 0.6978 & 1.5675 \\
MERGE~\cite{ganguly2025merge}
& 0.1699 & 0.1618 & 1.5830 & 1.0489 & \underline{0.8077}
&& 0.1750 & 0.1656 & 1.9248 & 1.1878 & 0.8823
&& 0.0672 & 0.0558 & 0.8466 & 0.7431 & 0.9111 \\
STFlow~\cite{huang2025stflow}
& \underline{0.3888} & 0.3410 & 0.8464 & \underline{0.7252} & 0.8555
&& \underline{0.5137} & \underline{0.4832} & 1.1983 & 0.8654 & \underline{0.4951}
&& \underline{0.1971} & \underline{0.1565} & \underline{0.7404} & \underline{0.6897} & \textbf{0.7245} \\
\midrule
\textbf{BCARP (Ours)}
& \textbf{0.4567} & \textbf{0.4032} & \textbf{0.8381} & \textbf{0.7191} & \textbf{0.7084}
&& \textbf{0.5852} & \textbf{0.5328} & 1.1818 & 0.8620 & \textbf{0.2656}
&& \textbf{0.3046} & \textbf{0.2423} & \textbf{0.6761} & \textbf{0.6590} & \underline{0.7500} \\
\bottomrule
\end{tabular}}
\end{table*}

\section{Implementation Details}
\label{sec:supp_impl}

\subsection{Preprocessing and supervision targets}
For each spot we crop a $224\!\times\!224$ H\&E patch centred on its tissue
coordinate and extract a frozen UNI~\cite{chen2024uni} embedding $\mathbf{e}_i\!\in\!\mathbb{R}^{d_e}$
with $d_e\!=\!1024$. Two complementary expression targets are used: a spatially smoothed target $\mathbf{Y}^{\mathrm{proc}}$,
obtained with the two-factor SPCS smoothing used by MERGE~\cite{ganguly2025merge} and used here to
supervise the pattern branch and the raw-count target $\mathbf{Y}^{\mathrm{raw}}$,
retained for the calibration objectives. Spatial structure is encoded once per
section as a fixed $k$-nearest-neighbour graph
$\mathbf{A}_s=\mathrm{kNN}(\widetilde{\mathbf{P}};k)$ built on max-normalised
coordinates $\widetilde{\mathbf{P}}$; this adjacency is shared across all
refinement steps and only specifies the local propagation structure.

\subsection{Nuclei-derived biological supervision}
Auxiliary local targets are derived from H\&E nuclei analysis with
HoVer-Net~\cite{graham2019hovernet}, which detects nuclei and assigns each a cell-type pseudo-label.
For each spot $i$ we aggregate nuclei whose centroids fall within a fixed
radius $\rho$ (in pixels, measured on the same patch grid as the
$224\!\times\!224$ crops) of $\mathbf{p}_i$. Letting $n_{it}$ be the number of type-$t$
nuclei in this neighborhood and $T$ the number of categories, the
neighborhood-composition target and dominant cell-type target are
\begin{equation}
\mathbf{q}_{i}
=
\big[
\frac{
n_{it}
}{
\sum_{t'=1}^{T} n_{it'}+\epsilon
}
\big]_{t=1}^{T},
\qquad
c_i
=
\arg\max_{t}
n_{it}.
\label{eq52}
\end{equation}
These quantities are used only as training-time auxiliary supervision and are
not required at inference.

\subsection{Architecture and optimization}
BCARP is implemented in PyTorch. A lightweight MLP projects each frozen UNI~\cite{chen2024uni}
embedding from its $d_e\!=\!1024$-dimensional space into a $d=512$-dimensional
transcriptomics-oriented latent space, which serves as the histology anchor. The
anchor is refined for $R\!=\!5$ context-conditioned steps; each step applies
context-conditioned anchor modulation, two GraphSAGE local context encoders
(cell-type and neighborhood-composition, hidden width $d_r\!=\!256$) and
prototype-guided global propagation with $K_{\mathrm{ctx}}\!=\!50$ learnable
global tokens and $H\!=\!8$ attention heads (head dimension $d_h\!=\!64$). The
refined representation is decoded by the pattern-scale decoder using $S\!=\!6$
hierarchy tokens. HoVer-Net~\cite{graham2019hovernet} targets use radius
$\rho\!=\!128$ pixels and $T\!=\!6$
nuclei categories. The model is trained with AdamW (learning rate
$1\!\times\!10^{-4}$, weight decay $1\!\times\!10^{-2}$), batch size $1$ WSI, for
at most $100$ epochs, with a \texttt{ReduceLROnPlateau} schedule (factor $0.5$,
patience $7$) on a single RTX\,5070\,Ti GPU. The configuration is summarised in
Table~\ref{tab:supp_hparams}.

\begin{table}[!ht]
\centering
\caption{Hyperparameter configuration of BCARP.}
\label{tab:supp_hparams}
\renewcommand{\arraystretch}{1.15}
\begin{tabular}{l}
\toprule
\textbf{Foundation Backbone} \\
\quad Pathology foundation encoder: \texttt{UNI}~\cite{chen2024uni} \\
\quad Spot patch size: \texttt{224} \\
\quad Embedding dimension $d_e$: \texttt{1024} \\
\quad Refinement iterations $R$: \texttt{5} \\
\midrule
\textbf{Biology-guided Context Modeling} \\
\quad Spatial neighbors $k$: \texttt{8} \\
\quad HoVer-Net~\cite{graham2019hovernet} neighborhood radius $\rho$: \texttt{128 px} \\
\quad Cell-type categories $T$: \texttt{6} \\
\quad Local encoder: \texttt{GraphSAGE} \\
\quad Local hidden dimension $d_r$: \texttt{256} \\
\quad Global prototype tokens $K_{\mathrm{ctx}}$: \texttt{50} \\
\quad Attention heads $H$: \texttt{8} \\
\quad Head dimension $d_h$: \texttt{64} \\
\midrule
\textbf{Decoder} \\
\quad Hierarchical decoder scales $S$: \texttt{6} \\
\quad Dropout: \texttt{0.1} \\
\midrule
\textbf{Training} \\
\quad Optimizer: AdamW \\
\quad Learning rate: $1\times10^{-4}$ \\
\quad Weight decay: $1\times10^{-2}$ \\
\quad Batch size: \texttt{1 WSI} \\
\quad Epochs: \texttt{100 (max)} \\
\quad Learning-rate scheduler: \texttt{ReduceLROnPlateau} \\
\quad \quad \texttt{(factor 0.5, patience 7)} \\
\quad Hardware: \texttt{RTX\,5070\,Ti} \\
\midrule
\textbf{Loss} \\
\quad Huber threshold $\delta$: \texttt{1.0} \\
\quad $\lambda_{\mathrm{Huber}}$: \texttt{0.5} \\
\quad $\lambda_{\mathrm{NB}}$: \texttt{0.05} \\
\quad $\lambda_{\mathrm{logMSE}}$: \texttt{0.1} \\
\quad $\lambda_{\mathrm{ct}}$: \texttt{0.05} \\
\quad $\lambda_{\mathrm{nc}}$: \texttt{0.05} \\
\bottomrule
\end{tabular}
\end{table}

\begin{table}[!ht]
\centering
\small
\caption{Impact of the nuclei-analysis backend on HER2ST. All other settings are
identical; only the pseudo-label generator changes.}
\label{tab:segmentation_ablation}
\resizebox{.475\textwidth}{!}{
\begin{tabular}{lccccc}
\toprule
\textbf{Labels} & \textbf{PCC-H}$\uparrow$ & \textbf{PCC-M}$\uparrow$ & \textbf{MSE}$\downarrow$ & \textbf{MAE}$\downarrow$ & \textbf{RVD}$\downarrow$ \\
\midrule
HoVer-Net~\cite{graham2019hovernet}
& 0.4567 & 0.4032 & \textbf{0.8381} & \textbf{0.7191} & 0.7084 \\
CellViT~\cite{CellViT}
& \textbf{0.4634} & \textbf{0.4060} & 0.8769 & 0.7426 & \textbf{0.6998} \\
\bottomrule
\end{tabular}}
\end{table}

\begin{table}[!ht]
\centering
\small
\caption{Computational cost on a single RTX\,5070\,Ti.}
\label{tab:supp_cost}
\resizebox{.475\textwidth}{!}{
\begin{tabular}{lrrr}
\toprule
\textbf{Method / component} & \textbf{Trainable (M)} & \textbf{E2E / slide (s)} & \textbf{Peak (GB)} \\
\midrule
\multicolumn{4}{l}{\emph{Baselines based on UNI embedding}} \\
UNI+Ridge                            &  0.26 & 1.28 & 0.22 \\
mclSTExp~\cite{min2024mclSTExp}  & 41.46 & 1.29 & 0.75 \\
MERGE~\cite{ganguly2025merge}    & 21.53 & 1.29 & 0.84 \\
STFlow~\cite{huang2025stflow}    &  1.35 & 1.32 & 1.03 \\
TRIPLEX~\cite{chung2024accurate} & 63.12 & 1.57 & 6.88 \\
EGN~\cite{yang2024exemplar}      & 77.01 & 1.59 & 3.80 \\
\textbf{BCARP (ours)}            & 21.10 & 1.37 & 1.35 \\
\midrule
\multicolumn{4}{l}{\emph{Native image encoder (no UNI)}} \\
M2OST~\cite{Wang2025m2ost}       &  7.04 & 0.85 & 11.75 \\
\bottomrule
\end{tabular}}
\end{table}

\paragraph{Foundation model controls.}
The HEST ridge-probe comparison uses ResNet50~\cite{He2016}, UNI~\cite{chen2024uni}, Virchow2~\cite{zimmermann2024virchow2},
CONCH~\cite{lu2024conch}, and GigaPath~\cite{Xu2024} embeddings with the same linear head and fixed
regularisation setting. We also examined OmiCLIP/Loki~\cite{chen2025visual} and STORM~\cite{Gurarslan2026}.
Loki (built on OmiCLIP) uses a transcriptomic reference bank at prediction time
and therefore follows a different reference-assisted setting. We
attempted to reproduce STORM using its released implementation but
could not establish a reliable matched reproduction under our
protocol, so we do not report potentially configuration-dependent
numbers.

\subsection{Evaluation metrics}
\label{sec:supp_metrics}
Let $\mathbf{Y}^{\mathrm{raw}},\hat{\mathbf{Y}}^{\mathrm{raw}}\!\in\!\mathbb{R}^{N\times G}$ be the
ground-truth and predicted raw-count matrices over $N$ test spots and $G$ genes. Unless
otherwise noted, all metrics are computed after the $\log(1+x)$ transform; we denote the
transformed matrices by
\begin{equation}
  \mathbf{Y}^{\mathrm{log}}:=\log\!\big(1+\mathbf{Y}^{\mathrm{raw}}\big),\qquad
  \hat{\mathbf{Y}}^{\mathrm{log}}:=\log\!\big(1+\hat{\mathbf{Y}}^{\mathrm{raw}}\big),
\end{equation}
and write $\mathbf{Y}^{\mathrm{log}}_{ig}$ and $\hat{\mathbf{Y}}^{\mathrm{log}}_{ig}$ for $i$-th spot and
$g$-th gene.

For each held-out fold, metrics are computed using only the spots in
that fold. Dataset-level scores in the main tables are the arithmetic
mean of the corresponding fold-level scores. For PCC, we first
compute each gene's correlation within a fold and then average over
the valid genes in the corresponding gene set.

\noindent\textbf{Per-gene Pearson correlation (PCC-M, PCC-H).}
For each gene we compute the PCC across spots,
\begin{equation}
  \mathrm{PCC}_g=
\frac{\sum_i (\hat{\mathbf{Y}}^{\mathrm{log}}_{ig}-\bar{\hat{\mathbf{Y}}}^{\mathrm{log}}_{g})(\mathbf{Y}^{\mathrm{log}}_{ig}-\bar{\mathbf{Y}}^{\mathrm{log}}_{g})}
     {\sqrt{\sum_i (\hat{\mathbf{Y}}^{\mathrm{log}}_{ig}-\bar{\hat{\mathbf{Y}}}^{\mathrm{log}}_{g})^2}\,
      \sqrt{\sum_i (\mathbf{Y}^{\mathrm{log}}_{ig}-\bar{\mathbf{Y}}^{\mathrm{log}}_{g})^2}},
  \label{eq:supp_pcc}
\end{equation}
where $\bar{\hat{\mathbf{Y}}}^{\mathrm{log}}_{g}$ and $\bar{\mathbf{Y}}^{\mathrm{log}}_{g}$ are averaged across all spots. 
We report PCC-M as the mean of $\mathrm{PCC}_g$ over all valid
target genes ($G=250$ for HER2ST, Skin, and ST-Net), and PCC-H
as the mean over the 50-gene HVG subset for these three datasets.
For HEST-1k, PCC-M is averaged over the cohort-specific target
panel described in Sec.~B.

\noindent\textbf{Relative variance distance (RVD).}
Because a high PCC in log-space does not by itself guarantee faithful
expression recovery, we consider RVD, which measures how well the
prediction preserves the per-gene heterogeneity of the data:
\begin{equation}
  \mathrm{RVD}=\frac{1}{G}\sum_{g=1}^{G}
  \frac{\left(\mathrm{Var}(\hat{\mathbf{Y}}^{\mathrm{log}}_{:,g})-\mathrm{Var}(\mathbf{Y}^{\mathrm{log}}_{:,g})\right)^2}
       {\left(\mathrm{Var}(\mathbf{Y}^{\mathrm{log}}_{:,g})\right)^2},
  \label{eq:supp_rvd}
\end{equation}
where $\mathrm{Var}(\cdot)$ is the across-spot variance. Lower RVD indicates
that the predicted expression reproduces the spatial variance of the ground
truth more faithfully. For all correlation metrics higher is better; for MSE,
MAE and RVD lower is better.

\noindent\textbf{Absolute-error metrics.}
\begin{equation}
  \mathrm{MSE}=\frac{1}{NG}\sum_{i,g}\left(\hat{\mathbf{Y}}^{\mathrm{log}}_{ig}-\mathbf{Y}^{\mathrm{log}}_{ig}\right)^2,
  \label{eq:supp_mse}
\end{equation}
\begin{equation}
  \mathrm{MAE}=\frac{1}{NG}\sum_{i,g}\left\lvert \hat{\mathbf{Y}}^{\mathrm{log}}_{ig}-\mathbf{Y}^{\mathrm{log}}_{ig}\right\rvert .
  \label{eq:supp_mae}
\end{equation}


\begin{figure*}
    \centering
    \includegraphics[width=1\linewidth]{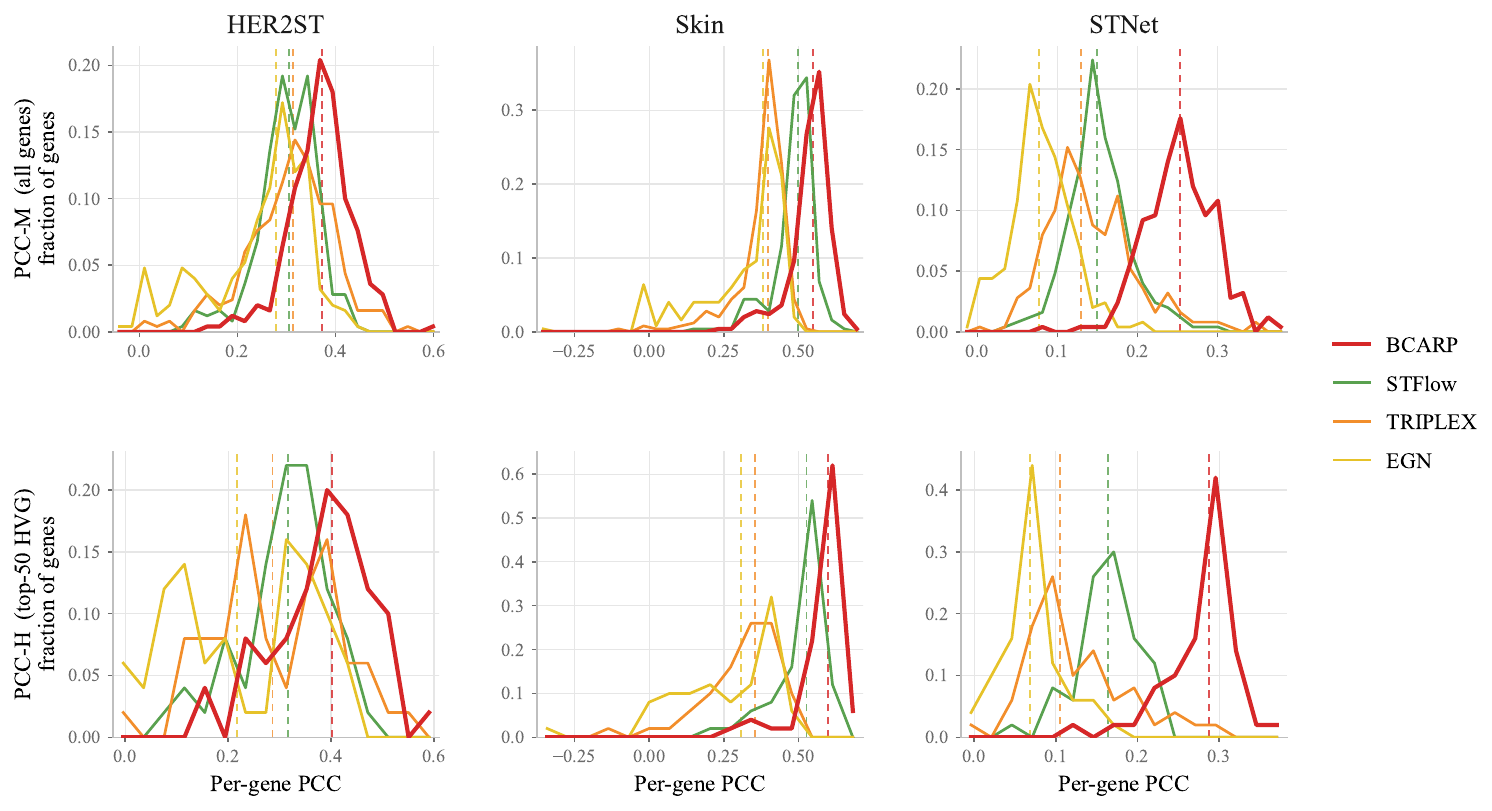}
    \caption{
Distribution of fold-averaged per-gene Pearson correlation coefficients
(PCC-H and PCC-M) across HER2ST~\cite{andersson2021spatial},
ST-Net~\cite{he2020integrating} and Skin~\cite{ji2020multimodal}. 
BCARP generally shifts the per-gene correlation distributions toward
higher values across both the full target panel and the HVG subset.
}
    \label{fig:pcc_hvg}
\end{figure*}

\begin{figure*}[!ht]
    \centering
    \includegraphics[width=1\linewidth]{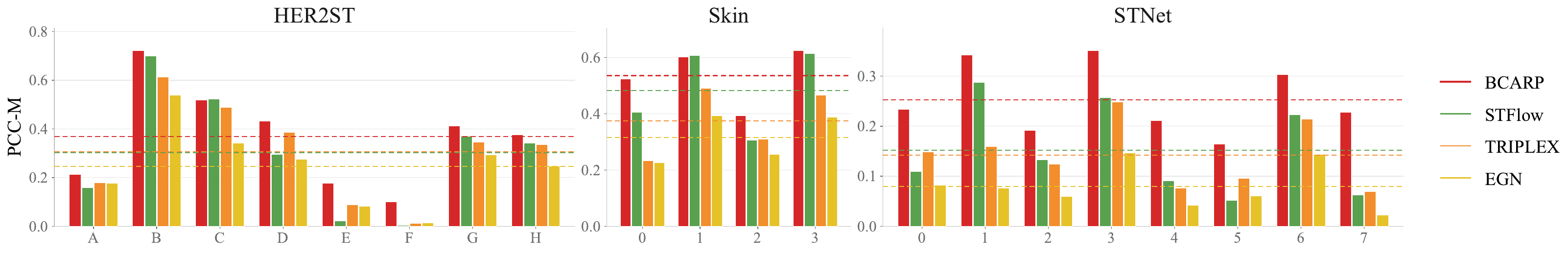}
    \caption{Per-fold PCC-M performance across the three ST datasets.}
    \label{fig:fold_pcc_st}
\end{figure*}

\begin{figure*}
    \centering
    \includegraphics[width=1\linewidth]{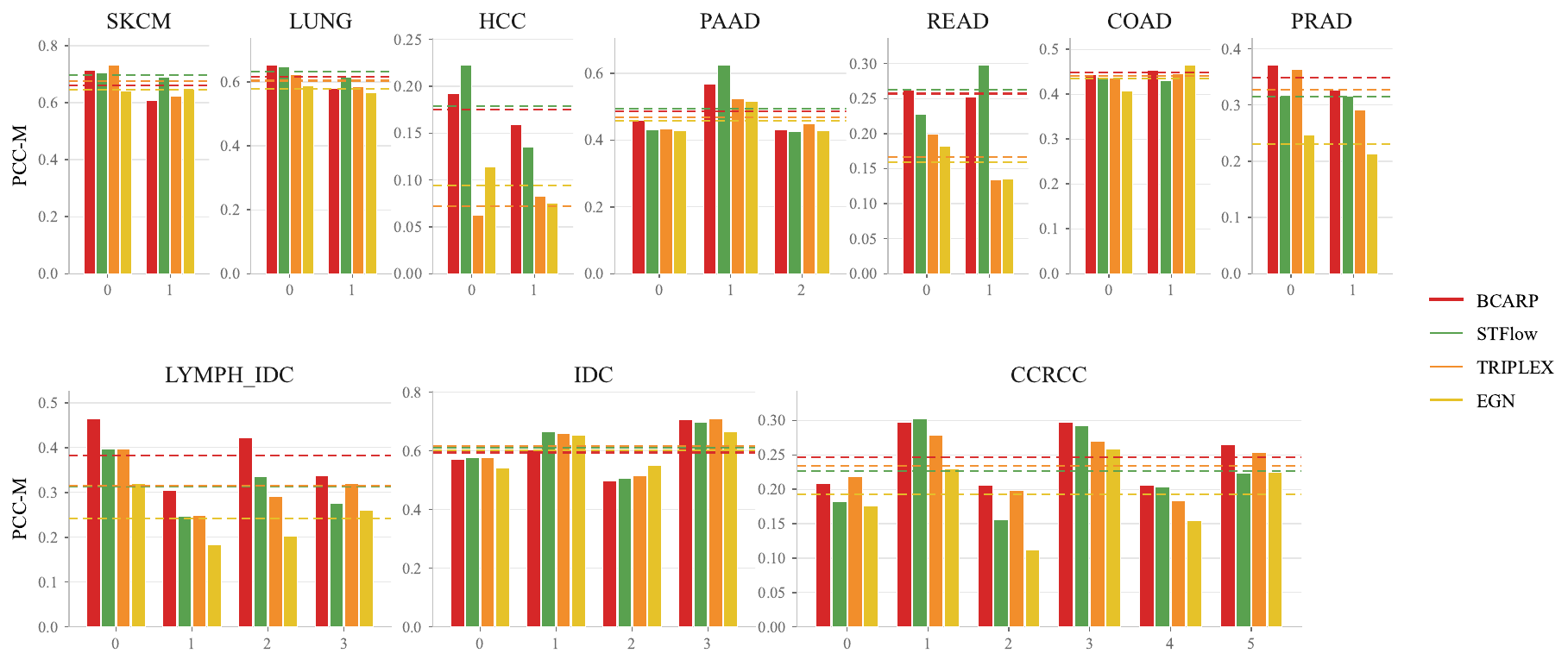}
    \caption{Per-fold PCC-M performance across the ten HEST-1k cohorts.
    Dashed lines mark each method's cohort mean.}
    \label{fig:fold_pcc_hest}
\end{figure*}

\begin{table*}[!ht]
\centering
\caption{Refinement-depth ablation as {relative change} (\%) of each metric with respect to the anchor $R\!=\!1$.}
\label{tab:refinement_ablation}
\resizebox{0.7\textwidth}{!}{
\begin{tabular}{llcccccccccc}
\toprule
\textbf{Dataset} & \textbf{Metric} & \textbf{1} & \textbf{2} & \textbf{3} & \textbf{4} & \textbf{5} & \textbf{6} & \textbf{7} & \textbf{8} & \textbf{9} & \textbf{10} \\
\midrule
\multirow{5}{*}{\textbf{HER2ST~\cite{andersson2021spatial}}}
 & PCC-H$\uparrow$ & +0.0 & +5.4 & +6.3 & +8.1 & +7.9 & +7.9 & +7.4 & \textbf{+8.5} & +8.1 & +7.6 \\
 & PCC-M$\uparrow$ & +0.0 & +5.7 & +6.1 & +7.8 & +7.5 & +7.5 & +7.4 & \textbf{+8.4} & +7.7 & +7.4 \\
 & MSE$\downarrow$  & +0.0 & -6.8 & -2.0 & -5.1 & -5.7 & -2.3 & -2.5 & \textbf{-7.4} & -3.9 & -3.8 \\
 & MAE$\downarrow$  & +0.0 & \textbf{-4.2} & -0.7 & -1.9 & -2.5 & -1.0 & -1.4 & -3.7 & -1.6 & -1.7 \\
 & RVD$\downarrow$ & +0.0 & \textbf{-14.3} & +18.4 & -12.6 & +1.7 & +7.1 & +11.6 & -11.7 & +7.9 & +10.2 \\
\midrule
\multirow{5}{*}{\textbf{Skin~\cite{ji2020multimodal}}}& PCC-H$\uparrow$ & +0.0 & +1.2 & +1.2 & +1.6 & +1.0 & +1.1 & +1.5 & \textbf{+1.8} & +1.1 & +1.1 \\
 & PCC-M$\uparrow$ & +0.0 & +0.7 & +0.6 & +0.7 & +0.2 & +0.2 & \textbf{+1.1} & +0.8 & +0.3 & +0.3 \\
 & MSE$\downarrow$  & +0.0 & \textbf{-5.5} & +24.8 & +29.4 & +29.0 & +28.2 & +0.8 & +31.9 & +27.3 & +29.4 \\
 & MAE$\downarrow$  & +0.0 & \textbf{-3.9} & +9.9 & +13.0 & +12.7 & +12.3 & -0.5 & +14.4 & +11.5 & +13.0 \\
 & RVD$\downarrow$ & +0.0 & +14.1 & \textbf{-8.7} & +9.8 & +3.5 & +11.7 & +12.5 & +15.5 & -0.2 & +10.2 \\
\midrule
\multirow{5}{*}{\textbf{ST-Net~\cite{he2020integrating}}}& PCC-H$\uparrow$ & +0.0 & -0.8 & -1.1 & -5.8 & \textbf{+0.8} & +0.7 & -2.2 & -3.6 & -3.6 & -0.8 \\
 & PCC-M$\uparrow$ & +0.0 & -1.2 & -1.4 & -4.9 & \textbf{+0.3} & +0.1 & -2.7 & -3.4 & -3.7 & -1.0 \\
 & MSE$\downarrow$  & +0.0 & \textbf{-6.2} & -4.4 & -3.1 & -2.8 & -1.0 & -4.9 & -0.8 & -3.5 & -3.3 \\
 & MAE$\downarrow$  & +0.0 & \textbf{-3.3} & -2.3 & -1.5 & -1.5 & -0.7 & -2.6 & -0.5 & -2.1 & -1.8 \\
 & RVD$\downarrow$ & +0.0 & -1.4 & -0.9 & +0.9 & \textbf{-4.4} & -2.7 & -0.2 & -0.3 & -2.8 & -0.8 \\
\midrule
\multirow{5}{*}{\textbf{Average}}& PCC-H$\uparrow$ & +0.0 & +2.0 & +2.1 & +1.3 & \textbf{+3.2} & \textbf{+3.2} & +2.2 & +2.2 & +1.9 & +2.7 \\
 & PCC-M$\uparrow$ & +0.0 & +1.8 & +1.8 & +1.2 & \textbf{+2.7} & +2.6 & +1.9 & +1.9 & +1.4 & +2.2 \\
 & MSE$\downarrow$  & +0.0 & \textbf{-6.2} & +6.1 & +7.1 & +6.8 & +8.3 & -2.2 & +7.9 & +6.6 & +7.4 \\
 & MAE$\downarrow$  & +0.0 & \textbf{-3.8} & +2.3 & +3.2 & +2.9 & +3.5 & -1.5 & +3.4 & +2.6 & +3.2 \\
 & RVD$\downarrow$ & +0.0 & \textbf{-0.6} & +2.9 & \textbf{-0.6} & +0.3 & +5.4 & +7.9 & +1.2 & +1.6 & +6.5 \\
\bottomrule
\end{tabular}
}
\end{table*} 

\begin{table*}[!ht]
\centering
\caption{Sensitivity to the number of global prototype tokens
$K_{\mathrm{ctx}}$ on HER2ST. Mean Imp. denotes the mean sign-corrected relative improvement across
the five metrics, where decreases in MSE, MAE, and RVD are counted as
positive improvements.}
\label{tab:kctx_ablation}
\resizebox{0.8\textwidth}{!}{
\begin{tabular}{lcccccc}
\toprule
$K_{\mathrm{ctx}}$ & \textbf{PCC-H$\uparrow$} & \textbf{PCC-M$\uparrow$} & \textbf{MSE$\downarrow$} & \textbf{MAE$\downarrow$} & \textbf{RVD$\downarrow$} & \textbf{Mean Imp.$\uparrow$} \\
\midrule
0
& \underline{0.4586}
& \underline{0.4061}
& \underline{0.8538}
& \underline{0.7282}
& 0.8123
& 0.00\%\\
5
& \textbf{0.4608} {\scriptsize($+$0.5\%)}
& \textbf{0.4076} {\scriptsize($+$0.4\%)}
& 0.8933 {\scriptsize($+$4.6\%)}
& 0.7507 {\scriptsize($+$3.1\%)}
& \underline{0.7522} {\scriptsize($-$7.4\%)}
& \underline{+0.11\%}\\
10
& 0.4509 {\scriptsize($-$1.7\%)}
& 0.3941 {\scriptsize($-$3.0\%)}
& 0.8807 {\scriptsize($+$3.2\%)}
& 0.7435 {\scriptsize($+$2.1\%)}
& 0.9480 {\scriptsize($+$16.7\%)}
& $-$5.32\%\\
25
& 0.4534 {\scriptsize($-$1.1\%)}
& 0.4043 {\scriptsize($-$0.4\%)}
& 0.9277 {\scriptsize($+$8.7\%)}
& 0.7657 {\scriptsize($+$5.1\%)}
& 0.8649 {\scriptsize($+$6.5\%)}
& $-$4.37\%\\
50
& 0.4567 {\scriptsize($-$0.4\%)}
& 0.4032 {\scriptsize($-$0.7\%)}
& \textbf{0.8381} {\scriptsize($-$1.8\%)}
& \textbf{0.7191} {\scriptsize($-$1.2\%)}
& \textbf{0.7084} {\scriptsize($-$12.8\%)}
& \textbf{+2.95\%}\\
100
& 0.4374 {\scriptsize($-$4.6\%)}
& 0.3891 {\scriptsize($-$4.2\%)}
& 0.8903 {\scriptsize($+$4.3\%)}
& 0.7537 {\scriptsize($+$3.5\%)}
& 0.7987 {\scriptsize($-$1.7\%)}
& $-$2.98\%\\
200
& 0.4530 {\scriptsize($-$1.2\%)}
& 0.3999 {\scriptsize($-$1.5\%)}
& 0.8591 {\scriptsize($+$0.6\%)}
& 0.7357 {\scriptsize($+$1.0\%)}
& 0.8464 {\scriptsize($+$4.2\%)}
& $-$1.72\%\\
\bottomrule
\end{tabular}}
\end{table*}

\section{Experimental Analysis}

\paragraph{Per-gene correlation distributions.}
Figure~\ref{fig:pcc_hvg} compares the fold-averaged per-gene PCC distributions
over the full panel and over the highly variable subset. Beyond shifting the bulk
of genes toward higher correlation, BCARP also trims the left tail. This is
visible on Skin~\cite{ji2020multimodal}, the only cohort whose distribution
extends below zero, where a non-negligible fraction of genes have PCC below zero
under TRIPLEX~\cite{chung2024accurate} and EGN~\cite{yang2024exemplar} while
almost all of BCARP's genes remain positive. STFlow is the closest competitor
throughout, and the gap to it is much smaller than the gap to TRIPLEX and EGN: on
Skin the two distributions overlap substantially and BCARP's advantage is a
modest rightward shift of the mode rather than a separation. The same ordering
holds on HER2ST~\cite{andersson2021spatial}, again with considerable overlap. The
separation is cleanest on ST-Net~\cite{he2020integrating}, where the BCARP
histogram is essentially disjoint from all three baselines, which concentrate
between $0.08$ and $0.16$ while BCARP peaks near $0.25$-$0.30$.

\paragraph{Per-fold stability.}
Figures~\ref{fig:fold_pcc_st} and~\ref{fig:fold_pcc_hest} report per-fold
PCC-M.
On the three ST datasets, BCARP is highest in every fold, and the
separation is most decisive on ST-Net~\cite{he2020integrating}, where it leads
at all eight folds, by more than a factor of two at folds~0 and~7 and by a
narrower margin elsewhere. The clearest exception is the
HER2ST~\cite{andersson2021spatial} folds corresponding to patients E and F
(folds~4-5), where all methods drop sharply and BCARP's margin narrows; this
points to specific tissue morphologies that remain difficult for every model
rather than to instability of BCARP itself. On the HEST-1k cohorts the picture
is more balanced and we report it as such: BCARP and STFlow trade the lead
fold by fold, with STFlow ahead on HCC, PAAD, READ~1 and LUNG~0, and BCARP
ahead on LYMPH\_IDC, PRAD and most CCRCC folds. On IDC all four methods are
within a narrow band and BCARP is not the strongest. 
BCARP's advantage on HEST-1k is therefore reflected in its aggregate
cross-cohort performance rather than uniform dominance within
individual cohorts, consistent with the mean-rank comparison in
Table~\ref{tab:supp_hparams} of the main paper.

\paragraph{Computational cost.}
Table~\ref{tab:supp_cost} reports three quantities measured under distinct conditions. Trainable parameters count only those updated by the optimiser and exclude the frozen encoder on which most methods depend. End-to-end latency covers the full inference path, encoder forward pass included, 
measured on a representative HER2ST section containing \(N=712\) spots.
Peak memory is the maximum allocator high-water mark observed over a single training epoch across the evaluated datasets.
For methods using frozen UNI embeddings, the shared encoder dominates
end-to-end latency. Against the 1.28\,s per-slide cost of UNI+Ridge,
BCARP adds only 0.09\,s, while the most expensive UNI-based heads add
approximately 0.3\,s. Since UNI embeddings can be extracted once and
cached, this shared cost is amortized during training.
M2OST does not use a separate frozen UNI encoder and therefore has the
lowest measured end-to-end latency, but it incurs the largest peak
memory footprint. BCARP remains within the same memory range as the
other embedding-based heads while training fewer parameters than
mclSTExp, TRIPLEX, and EGN.


\paragraph{Sensitivity to the number of global prototype tokens.}

Table~\ref{tab:kctx_ablation} varies $K_{\mathrm{ctx}}$, taking
$K_{\mathrm{ctx}}=0$, a prototype-free global stage that retains
spot-level self-attention, as the reference.
The response is not monotonic. Correlation
peaks at a very small number of tokens: $K_{\mathrm{ctx}}\!=\!5$ gives the best
PCC-H and PCC-M, and both are marginally above the prototype-free baseline,
which indicates that the prototype-guided component contributes only
modestly to correlation on HER2ST.
Its contribution is concentrated instead in variance fidelity,
where $K_{\mathrm{ctx}}\!=\!50$ reduces RVD by $12.8\%$ relative to the
baseline, the largest single \emph{improvement} in the table, while also being
the only setting that improves both absolute-error metrics. Aggregated over the
five metrics, $K_{\mathrm{ctx}}\!=\!50$ is the only configuration with a clearly
positive mean improvement ($+2.95\%$) and we adopt it as the default. We note
that four of the six non-zero token budgets score below the prototype-free
baseline, so the benefit of prototype-guided propagation is
token-budget dependent rather than monotonic in
$K_{\mathrm{ctx}}$.



\paragraph{Sensitivity to the nuclei-analysis backend.}
The neighborhood-composition and cell-type targets of Equation~\eqref{eq52}
depend on an off-the-shelf nuclei segmentation model. To check our
conclusions, we regenerate both targets with
CellViT~\cite{CellViT} and retrain BCARP on HER2ST
(Table~\ref{tab:segmentation_ablation}). All other settings are identical and
only the pseudo-label generator changes. The two sources give closely comparable
results, and the differences do not point uniformly in one direction: CellViT is
slightly better on the correlation metrics and on RVD, while
HoVer-Net~\cite{graham2019hovernet} is better on both absolute-error metrics. 
The CellViT experiment is therefore used as a
post-hoc robustness check rather than for model selection. Since the two
backends yield comparable downstream performance without a consistent
winner across metrics, we retain HoVer-Net for consistency with all main
experiments.
These results indicate that BCARP's HER2ST performance is not
highly sensitive to the choice between the two tested nuclei-analysis backends,
although broader cross-tissue robustness remains to be studied.

\begin{table*}[!ht]
\centering
\caption{Detailed cross-validation splits by fold. For HER2ST, Skin and ST-Net, test partitions are identified by patient; all sections from a patient are assigned exclusively to the training or the test partition. For the HEST-1k cohorts we adopt the official patient-grouped folds released with the benchmark and identify test partitions by HEST slide ID.}
\label{tab:supp_splits}
\footnotesize
\setlength{\tabcolsep}{5pt}
\begin{tabular}{llcccc}
\toprule
Dataset & Fold & Test partition & Train slides & Test slides & Train spots / Test spots \\
\midrule
\multicolumn{6}{l}{\emph{ST datasets}} \\
\midrule
\multirow{8}{*}{HER2ST~\cite{andersson2021spatial}}
 & 0 & A & 30 & 6 & 11{,}555 / 2{,}065 \\
 & 1 & B & 30 & 6 & 11{,}908 / 1{,}712 \\
 & 2 & C & 30 & 6 & 12{,}534 / 1{,}086 \\
 & 3 & D & 30 & 6 & 11{,}787 / 1{,}833 \\
 & 4 & E & 33 & 3 & 11{,}891 / 1{,}729 \\
 & 5 & F & 33 & 3 & 11{,}522 / 2{,}098 \\
 & 6 & G & 33 & 3 & 12{,}249 / 1{,}371 \\
 & 7 & H & 33 & 3 & 11{,}894 / 1{,}726 \\
\midrule
\multirow{4}{*}{Skin~\cite{ji2020multimodal}}
 & 0 & P10\_rep & 9 & 3 & 6{,}980 / 1{,}691 \\
 & 1 & P2\_rep  & 9 & 3 & 6{,}721 / 1{,}950 \\
 & 2 & P5\_rep  & 9 & 3 & 7{,}039 / 1{,}632 \\
 & 3 & P9\_rep  & 9 & 3 & 5{,}273 / 3{,}398 \\
\midrule
\multirow{8}{*}{ST-Net~\cite{he2020integrating}}
 & 0 & BC23209\,C/D, BC23377\,C/D, BC23901\,C/D & 60 & 8 & 27{,}215 / 3{,}412 \\
 & 1 & BC23268\,C/D, BC23450\,D/E, BC23903\,C/D & 59 & 9 & 26{,}896 / 3{,}731 \\
 & 2 & BC23269\,C/D, BC23506\,C/D, BC23944\,D/E & 59 & 9 & 26{,}616 / 4{,}011 \\
 & 3 & BC23270\,D/E, BC23508\,D/E, BC24044\,D/E & 59 & 9 & 26{,}598 / 4{,}029 \\
 & 4 & BC23272\,D/E, BC23567\,D/E, BC24105\,C/D & 59 & 9 & 26{,}248 / 4{,}379 \\
 & 5 & BC23277\,D/E, BC23803\,D/E, BC24220\,D/E & 59 & 9 & 26{,}587 / 4{,}040 \\
 & 6 & BC23287\,C/D, BC23810\,D/E, BC24223\,D/E & 59 & 9 & 26{,}490 / 4{,}137 \\
 & 7 & BC23288\,D/E, BC23895\,C/D & 62 & 6 & 27{,}739 / 2{,}888 \\
\midrule
\multicolumn{6}{l}{\emph{HEST-1k cohorts}} \\
\midrule
\multirow{2}{*}{SKCM}
 & 0 & TENX117 & 1 & 1 & 1{,}741 / 1{,}293 \\
 & 1 & TENX115 & 1 & 1 & 1{,}293 / 1{,}741 \\
\midrule
\multirow{2}{*}{LUNG}
 & 0 & TENX141 & 1 & 1 & 1{,}944 / 3{,}262 \\
 & 1 & TENX118 & 1 & 1 & 3{,}262 / 1{,}944 \\
\midrule
\multirow{3}{*}{PAAD}
 & 0 & TENX116 & 2 & 1 & 5{,}561 / 2{,}010 \\
 & 1 & TENX140 & 2 & 1 & 3{,}961 / 3{,}610 \\
 & 2 & TENX126 & 2 & 1 & 5{,}620 / 1{,}951 \\
\midrule
\multirow{2}{*}{READ}
 & 0 & ZEN48, ZEN49 & 2 & 2 & 3{,}819 / 4{,}588 \\
 & 1 & ZEN36, ZEN40 & 2 & 2 & 4{,}588 / 3{,}819 \\
\midrule
\multirow{2}{*}{COAD}
 & 0 & TENX147-TENX149 & 1 & 3 & 5{,}079 / 10{,}572 \\
 & 1 & TENX111 & 3 & 1 & 10{,}572 / 5{,}079 \\
\midrule
\multirow{4}{*}{LYMPH\_IDC}
 & 0 & NCBI684 & 3 & 1 & 14{,}972 / 4{,}992 \\
 & 1 & NCBI683 & 3 & 1 & 14{,}972 / 4{,}992 \\
 & 2 & NCBI682 & 3 & 1 & 14{,}976 / 4{,}988 \\
 & 3 & NCBI681 & 3 & 1 & 14{,}972 / 4{,}992 \\
\midrule
\multirow{6}{*}{CCRCC}
 & 0 & INT13-INT15, INT24 & 20 & 4 & 55{,}849 / 18{,}371 \\
 & 1 & INT9-INT12 & 20 & 4 & 67{,}651 / 6{,}569 \\
 & 2 & INT6-INT8, INT23 & 20 & 4 & 63{,}464 / 10{,}756 \\
 & 3 & INT2-INT5 & 20 & 4 & 67{,}098 / 7{,}122 \\
 & 4 & INT1, INT20-INT22 & 20 & 4 & 59{,}472 / 14{,}748 \\
 & 5 & INT16-INT19 & 20 & 4 & 57{,}566 / 16{,}654 \\
\midrule
\multirow{4}{*}{IDC}
 & 0 & TENX99 & 3 & 1 & 14{,}775 / 20{,}761 \\
 & 1 & TENX95 & 3 & 1 & 27{,}776 / 7{,}760 \\
 & 2 & NCBI785 & 3 & 1 & 31{,}526 / 4{,}010 \\
 & 3 & NCBI783 & 3 & 1 & 32{,}531 / 3{,}005 \\
\midrule
\multirow{2}{*}{PRAD}
 & 0 & MEND154, MEND156-MEND162 & 15 & 8 & 36{,}819 / 25{,}891 \\
 & 1 & MEND139-MEND153 & 8 & 15 & 25{,}891 / 36{,}819 \\
\midrule
\multirow{2}{*}{HCC}
 & 0 & NCBI643 & 1 & 1 & 1{,}941 / 2{,}255 \\
 & 1 & NCBI642 & 1 & 1 & 2{,}255 / 1{,}941 \\
\bottomrule
\end{tabular}
\end{table*}

\begin{figure*}[!ht]
    \centering
    \includegraphics[width=1\linewidth]{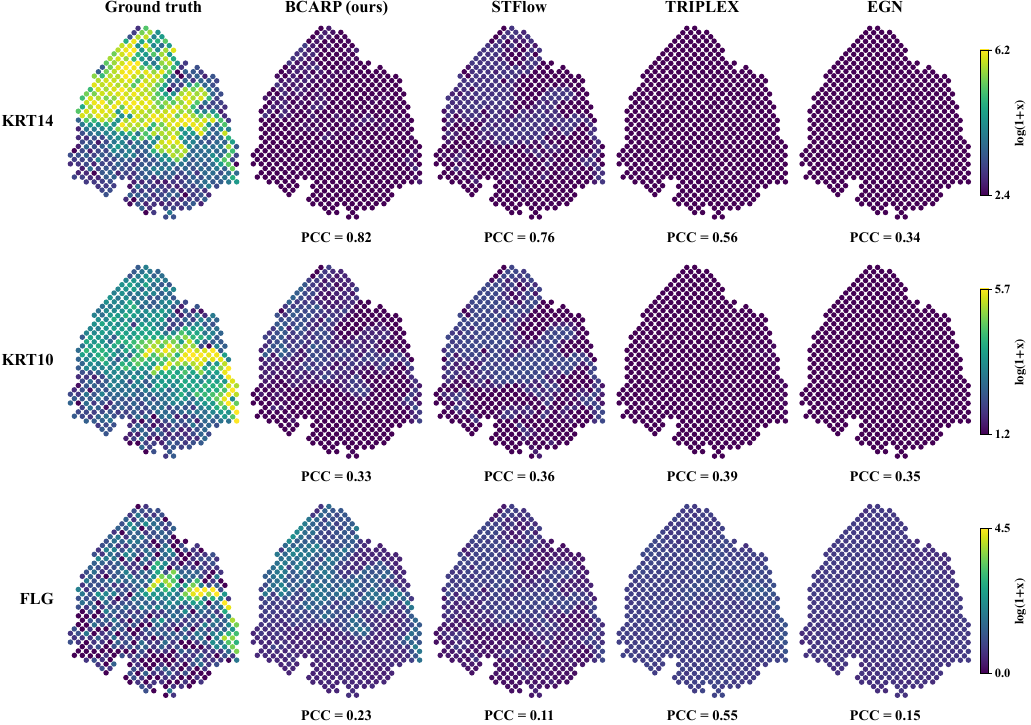}
    \caption{Qualitative comparison of predicted expression for three marker genes on Skin sample P2\_rep3.}
\end{figure*}

\begin{figure*}[!ht]
    \centering
    \includegraphics[width=0.9\linewidth]{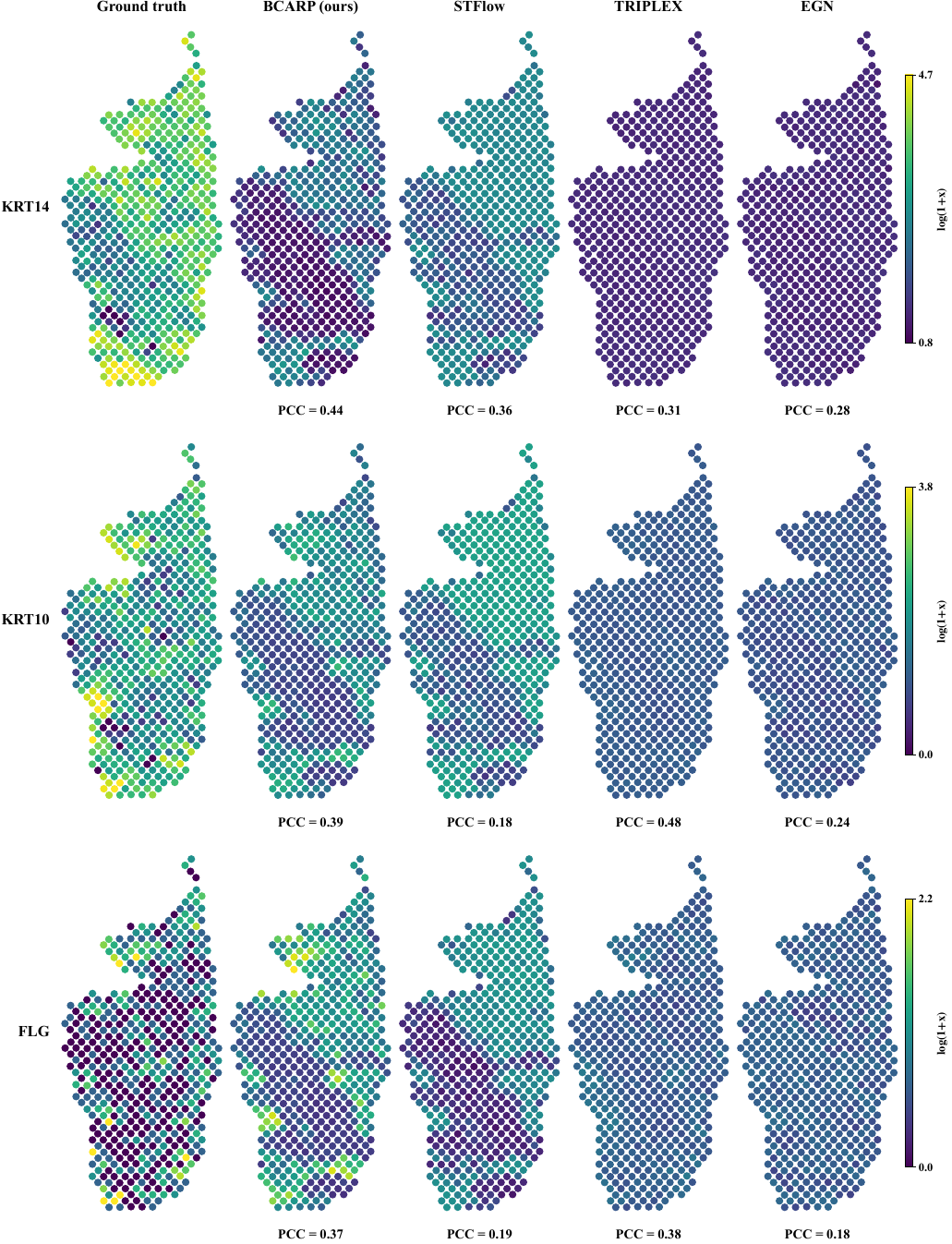}
    \caption{Qualitative comparison of predicted expression for three marker genes on Skin sample P5\_rep2.}
\end{figure*}

\begin{figure*}[!ht]
    \centering
    \includegraphics[width=1\linewidth]{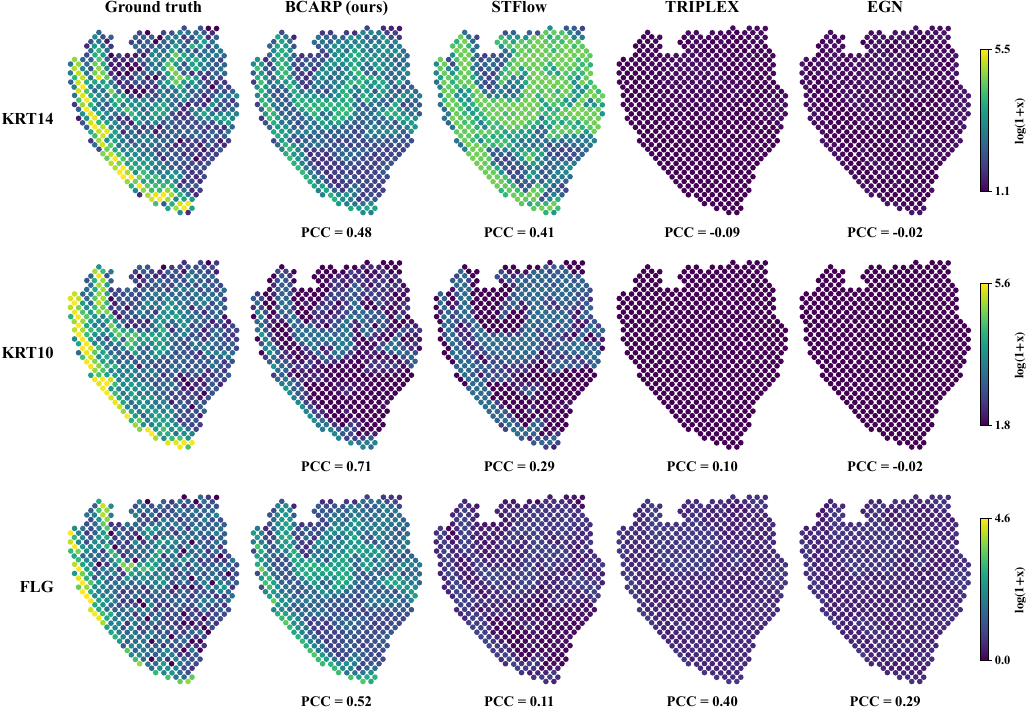}
    \caption{Qualitative comparison of predicted expression for three marker genes on Skin sample P10\_rep2}
\end{figure*}


\end{document}